\documentclass[conference]{IEEEtran}
\IEEEoverridecommandlockouts
\usepackage{amsmath,amssymb,amsfonts}
\usepackage[ruled, lined, linesnumbered, longend]{algorithm2e}
\usepackage[accsupp]{axessibility}
\usepackage{booktabs}
\usepackage{cite}
\usepackage{float}
\usepackage{graphicx}
\usepackage{siunitx}
\usepackage[caption=false]{subfig}
\usepackage{textcomp}
\usepackage{url}
\usepackage[dvipsnames]{xcolor}
\usepackage{xspace}

\usepackage{pgfplots}
\usepackage{tikz}
\DeclareUnicodeCharacter{2212}{−}
\usepgfplotslibrary{groupplots,dateplot}
\usetikzlibrary{patterns, shapes.geometric, shapes.arrows}
\pgfplotsset{compat=newest}

\makeatletter
\DeclareRobustCommand\onedot{\futurelet\@let@token\@onedot}
\def\@onedot{\ifx\@let@token.\else.\null\fi\xspace}

\def\eg{\emph{e.g}\onedot} 
\def\ie{\emph{i.e}\onedot}

\def\etal{\emph{et~al}\onedot}
\makeatother

\newlength{\arrowsize}  
\pgfarrowsdeclare{biggertip}{biggertip}{  
  \setlength{\arrowsize}{1pt}  
  \addtolength{\arrowsize}{.5\pgflinewidth}  
  \pgfarrowsrightextend{0}  
  \pgfarrowsleftextend{-5\arrowsize}  
}{  
  \setlength{\arrowsize}{1pt}  
  \addtolength{\arrowsize}{.5\pgflinewidth}  
  \pgfpathmoveto{\pgfpoint{-5\arrowsize}{4\arrowsize}}  
  \pgfpathlineto{\pgfpointorigin}  
  \pgfpathlineto{\pgfpoint{-5\arrowsize}{-4\arrowsize}}  
  \pgfusepathqstroke  
}

\SetCommentSty{mycommfont}

\tikzstyle{convlayer} = [
    rectangle,
    rounded corners, 
    minimum width=1.5cm, 
    minimum height=0.8cm,
    text centered, 
    draw=black
]

\tikzstyle{aux_head} = [
    rectangle,
    rounded corners, 
    dashed,
    minimum width=6.8cm, 
    minimum height=1.6cm,
    draw=black
]

\tikzstyle{conv} = [
    circle, 
    text centered,
    draw=black
]

\newcommand\copyrighttext{\footnotesize \textcopyright~2023 IEEE. Personal use of this material is permitted. Permission from IEEE must be obtained for all other uses, in any current or future media, including reprinting/republishing this material for advertising or promotional purposes, creating new collective works, for resale or redistribution to servers or lists, or reuse of any copyrighted component of this work in other works.
DOI: \href{https://doi.org/10.1109/IV55152.2023.10186625}{10.1109/IV55152.2023.10186625}
}

\newcommand\copyrightnotice{%
    \begin{tikzpicture}[remember picture,overlay]%
     \node[anchor=south, xshift=0pt, yshift=20pt] at (current page.south)%
     {\fbox{\parbox{\dimexpr\textwidth-\fboxsep-\fboxrule\relax}{\copyrighttext}}};%
     \end{tikzpicture}%
}

\def\BibTeX{{\rm B\kern-.05em{\sc i\kern-.025em b}\kern-.08em
    T\kern-.1667em\lower.7ex\hbox{E}\kern-.125emX}}
\begin{document}

\title{RT-K-Net: Revisiting K-Net for \\ Real-Time Panoptic Segmentation
\thanks{This research is accomplished within the project UNICARagil (FKZ 16EMO0290). We acknowledge the financial support for the project by the Federal Ministry of Education and Research of Germany (BMBF).}
}

\author{\IEEEauthorblockN{Markus Schön, Michael Buchholz, and Klaus Dietmayer}
\IEEEauthorblockA{\textit{Institute of Measurement, Control and Microtechnology} \\
\textit{Ulm University}, Germany \\
\{\tt\small markus.schoen,michael.buchholz,klaus.dietmayer\}@uni-ulm.de}
}

\maketitle

\begin{abstract}
Panoptic segmentation is one of the most challenging scene parsing tasks, combining the tasks of semantic segmentation and instance segmentation.
While much progress has been made, few works focus on the real-time application of panoptic segmentation methods.
In this paper, we revisit the recently introduced K-Net architecture.
We propose vital changes to the architecture, training, and inference procedure, which massively decrease latency and improve performance.
Our resulting RT-K-Net sets a new state-of-the-art performance for real-time panoptic segmentation methods on the Cityscapes dataset and shows promising results on the challenging Mapillary Vistas dataset.
On Cityscapes, RT-K-Net reaches 60.2~\%~PQ with an average inference time of 32~ms for full resolution $\text{1024} \times \text{2048}$ pixel images on a single Titan RTX GPU.
On Mapillary Vistas, RT-K-Net reaches 33.2~\%~PQ with an average inference time of 69~ms.
Source code is available at \url{https://github.com/markusschoen/RT-K-Net}.

\end{abstract}
\section{Introduction}
\copyrightnotice
Scene understanding is essential in automated driving perception systems since higher-order functions such as behavior planning need detailed information about the vehicle's environment to make reasonable decisions.
Camera-based scene understanding often boils down to image segmentation, \ie, finding meaningful groups of pixels in an image.
\begin{figure}[t!]
\begin{center}
\begin{tikzpicture}

\pgfplotsset{every tick label/.append style={font=\scriptsize}}
\definecolor{darkgray176}{RGB}{176,176,176}
\definecolor{darkviolet1910191}{RGB}{191,0,191}
\definecolor{green01270}{RGB}{0,127,0}
\definecolor{lightgray204}{RGB}{204,204,204}

\begin{axis}[
legend cell align={left},
legend style={fill opacity=0.8, draw opacity=1, text opacity=1, draw=lightgray204, font=\scriptsize},
tick align=outside,
tick pos=left,
x grid style={darkgray176},
xlabel={\scriptsize Inference Speed in ms},
xmajorgrids,
xmin=20, xmax=120,
xtick style={color=black},
y grid style={darkgray176},
ylabel={\scriptsize Panoptic Quality in \%},
ymajorgrids,
ymin=55, ymax=62,
ytick style={color=black}
]
\addplot [semithick, red, mark=+, mark size=3, mark options={solid}, only marks]
table {%
32 60.2
};
\addlegendentry{RT-K-Net~(Ours)}
\addplot [semithick, red, mark=asterisk, mark size=3, mark options={solid}, only marks]
table {%
112 56.9
};
\addlegendentry{K-Net~\cite{Zhang2021}}
\addplot [semithick, blue, mark=star, mark size=3, mark options={solid}, only marks]
table {%
36 55.7
};
\addlegendentry{MGNet~\cite{Schoen2021}}
\addplot [semithick, black, mark=square*, mark size=3, mark options={solid}, only marks]
table {%
63 58.4
};
\addlegendentry{Panoptic Deeplab~\cite{Chen2020}}
\addplot [semithick, green01270, mark=pentagon*, mark size=3, mark options={solid}, only marks]
table {%
82 57.3
};
\addlegendentry{Petrovai and Nedevschi~\cite{Petrovai2020}}
\addplot [semithick, darkviolet1910191, mark=diamond*, mark size=3, mark options={solid}, only marks]
table {%
99 58.8
};
\addlegendentry{Hou~\etal~\cite{Hou2020}}
\draw[-biggertip,draw=red] (axis cs:110,57) -- (axis cs:34,60.1);
\end{axis}

\end{tikzpicture}
\end{center}
\vspace{-6mm}
   \caption{Comparison of real-time state-of-the-art panoptic segmentation methods on the Cityscapes~\cite{Cordts2016} validation set at $1024 \times 2048$ resolution. Our method achieves the best speed-accuracy trade-off.}
\label{fig:infer_time}
\vspace{-3mm}
\end{figure}
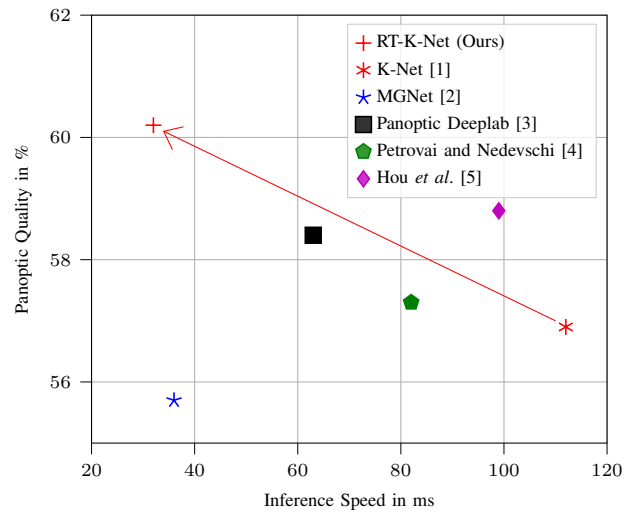
\mbox{Kirillov~\etal~\cite{Kirillov2019}} introduced panoptic segmentation, which yields a more complete scene understanding.
In panoptic segmentation, classes are divided into stuff classes, \ie, amorph regions such as road or sky, and thing classes, \ie, countable objects such as cars or pedestrians.
Panoptic segmentation requires not only grouping pixels by their class as in semantic segmentation but also grouping pixels of thing classes by object instances as in instance segmentation.
Hence, panoptic segmentation provides detailed information relevant to real-world applications such as automated driving, \eg, it gives information about other traffic participants as well as the current drivable space of the vehicle.
Research in panoptic segmentation often focuses on high accuracy~\cite{Cheng2021_1, Li2021_2, Mohan2021, Porzi2019, Yu2022} rather than fast inference speed.
While early methods~\cite{Mohan2021, Porzi2019} used separate formulations for semantic segmentation and instance segmentation, recent state-of-the-art methods~\cite{Zhang2021, Cheng2021_1, Yu2022} use a unified mask classification approach for panoptic segmentation.
The unified formulation reduces training complexity and achieves higher accuracy than specialized formulations.
However, both high accuracy and fast inference speed are necessary for real-world applications, and methods such as Mask2Former~\cite{Cheng2021_1} rely on heavy architectures, making them unsuitable for real-time applications.
Existing real-time panoptic segmentation methods~\cite{Petrovai2020, Hou2020, Schoen2021, Chen2020} still rely on specialized formulations for semantic segmentation and instance segmentation.
Therefore, these methods lack behind in accuracy compared to unified state-of-the-art methods.

This paper introduces RT-K-Net to close the gap between existing real-time and recent unified panoptic segmentation methods.
RT-K-Net is based on K-Net~\cite{Zhang2021}, a unified panoptic segmentation method using learned kernels to predict masks for both thing and stuff classes.
K-Net is lightweight compared to Mask2Former but still has several shortcomings concerning inference speed and overall performance for real-time applications.
Firstly, we find that K-Net cannot run with mixed precision due to numeric overflow in the kernel update modules.
Mixed precision is a simple and effective technique to increase inference speed without sacrificing performance.
Hence, we develop a simple but effective mask normalization technique to keep values within the half-precision range.
Secondly, we optimize the post-processing module of \mbox{K-Net} using element-wise multiplication of predicted labels and masks to parallelize the mask pasting step.
Thirdly, we make further changes to the K-Net architecture,~\ie, using the recently introduced \mbox{RTFormer~\cite{Wang2022}} as Backbone and Neck, and directly using panoptic kernels with an auxiliary semantic head in the initialization stage.
Finally, we improve the instance segmentation quality by introducing an instance-aware crop augmentation and an instance discrimination loss~\cite{Yu2022_1}.
Our resulting \mbox{RT-K-Net} is the first unified real-time panoptic segmentation method.
Fig.~\ref{fig:infer_time} shows a comparison between RT-K-Net and other real-time methods on the Cityscapes dataset~\cite{Cordts2016}. 
\mbox{RT-K-Net} outperforms all other methods in the PQ metric while also being the fastest overall.
In summary, our contributions are the following:
\begin{itemize}
\item 
We develop a simple and effective mask normalization technique, which improves group feature assembling and enables mixed precision training and inference for K-Net.
\item
We improve the inference speed of post-processing without sacrificing performance by efficiently combining the mask predictions with the label predictions.
Any recent unified panoptic segmentation architecture, such as~\cite{Yu2022, Cheng2021_1, Zhang2021}, can use this optimized version.
\item
We adjust the K-Net feature map generation and mask initialization stages and improve the instance discrimination ability of the model by using an instance-aware crop augmentation and a contrastive loss.
\item
Our resulting RT-K-Net outperforms all previous real-time methods on the Cityscapes panoptic segmentation benchmark.
In particular, our model reaches $60.2\%$~PQ with an average inference time of $32~\si{ms}$ for full resolution $1024 \times 2048$ pixel images on a single Titan RTX GPU.
\end{itemize}

\section{Related Work}
Panoptic segmentation~\cite{Kirillov2019} was introduced to unify the tasks of semantic and instance segmentation.
Early works can be divided into top-down and bottom-up approaches based on their approach to instance segmentation.
Top-down approaches~\cite{Porzi2019, Mohan2021} first generate proposals, usually in the form of bounding boxes, which are then used to predict instances.
For example, Porzi~\etal~\cite{Porzi2019} use a Mask-RCNN~\cite{He2017} head for instance segmentation and a variant of DeepLabV3~\cite{Chen2017} for semantic segmentation and combine both predictions in a downstream fusion module.
In contrast, bottom-up approaches~\cite{Gao2019, Cheng2020} are proposal-free.
For example, Panoptic-DeepLab~\cite{Cheng2020} represents instance masks as pixel offsets and center keypoints and fuses them with semantic logits to generate panoptic segmentation.
Recently, unified architectures~\cite{Li2021_2, Cheng2021, Cheng2021_1, Yu2022, Yu2022_1, Zhang2021} emerged.
These methods use a mask classification approach~\cite{Cheng2021} to predict segmentation masks with corresponding classes directly.
For example, K-Net~\cite{Zhang2021} uses learned convolutional kernels and an iterative kernel update to generate discriminative kernels.
The kernels are then convolved with a feature map to generate mask predictions. 
RT-K-Net is based on K-Net but addresses several shortcomings to improve performance and inference speed for real-time applications.

Only a few works exist focusing on real-time panoptic segmentation~\cite{Petrovai2020, Hou2020, Chen2020, Schoen2021}.
For example, Chen~\etal~\cite{Chen2020} push the performance of Panoptic-DeepLab~\cite{Cheng2020} by replacing the network's backbone with Scaling Wide Residual Networks (SWideRNets).
While such methods provide fast inference speeds, they lack accuracy compared to state-of-the-art methods due to separate semantic and instance formulations.
Our RT-K-Net is the first unified architecture for real-time panoptic segmentation.
\section{Method}
\label{sec:method}

In this section, we describe our RT-K-Net architecture for real-time panoptic segmentation.
Since our method is based on K-Net~\cite{Zhang2021}, we first revisit K-Net, before explaining our changes for RT-K-Net.
Finally, we describe two methods to increase the instance segmentation performance of \mbox{RT-K-Net}.

\subsection{K-Net Architecture}
\label{sec:KNet}
The core idea of K-Net is to learn convolutional kernels, which discriminate pixels into $N$ meaningful groups.
A meaningful group can either be an amorph region or a single object, enabling semantic segmentation, instance segmentation, and panoptic segmentation using the same formulation.
Given a feature map $F\in\mathbb{R}^{B\times C \times H \times W}$ and $N$ kernels $K\in\mathbb{R}^{B\times N \times C}$, K-Net produces $N$ mask predictions by performing convolution of $F$ and $K$
\begin{equation}
    M = K \ast F
\end{equation}
with $M\in\mathbb{R}^{B\times N \times H \times W}$.

K-Net uses $N$ randomly initialized Kernels $K_0$ to produce an initial mask prediction $M_0$.
Since initial kernels $K_0$ are not discriminative enough, especially for instance segmentation, K-Net performs an iterative kernel update, which uses the initial mask prediction $M_0$ and feature map $F$ to update the kernels.
This kernel update is performed in 3 steps:
\begin{enumerate}
\item 
\textbf{Group Feature Assembling:} 
Given the mask prediction $M_{i-1}$ from the last update stage (or the initial mask prediction $M_0$), group features are formed using element-wise multiplication of the binarized mask $S_{i-1}$ and feature map $F$
\begin{subequations}
\begin{gather}
	F^{K} = \sum_{u}^{H}\sum_{v}^{W} S_{i-1}(u, v) \cdot F(u, v), \\
    \text{with } S_{i-1}(u, v) = \sigma(M_{i-1}(u, v)) > 0.5.
\end{gather}
\end{subequations}
\item
\textbf{Adaptive Kernel Update:}
Since group features can contain noise, \eg, if the mask prediction contains pixels from different groups, kernels are updated using a weighted sum of initial kernels and group features
\begin{equation}
    \tilde{K} = {G}^{F} \otimes \psi_1(F^{K})  + {G}^{K} \otimes \psi_2(K_{i-1}).
\end{equation}
${G}^{F}$ and ${G}^{K}$ are learned gates to weight the influence of kernels and group features to the kernel update, whereas $\psi_1$ and $\psi_2$ are linear transformations.
\item
\textbf{Kernel Interaction:}
Kernel interaction is performed to model the relationship between kernels, \ie, between different pixel groups.
Here, K-Net adopts Multi-Head Self-Attention (MHSA)~\cite{Vaswani2017} followed by a Feed-Forward Neural Network (FFN)
\begin{equation}
    K_i = \mathrm{FFN}(\mathrm{MHSA}(\tilde{K})).
\end{equation}
\end{enumerate}
The updated kernels $K_i$ are used to generate a new mask prediction, and a class probability prediction
\begin{equation}
    M_i = \mathrm{FFN}_M(K_i) \ast F,\quad p_i = \mathrm{FFN}_p(K_i).
\end{equation}
During training, K-Net uses a fixed matching strategy for assigning stuff masks to their respective ground truth mask.
For thing masks, K-Net adopts bipartite matching using the Hungarian method and a set prediction loss similar to~\cite{Cheng2021}.
The final loss for K-Net consists of
\begin{equation}
    \mathcal{L} = \lambda_{\mathrm{mask}}\mathcal{L}^{\mathrm{mask}} + \lambda_{\mathrm{dice}}\mathcal{L}^{\mathrm{dice}} + \lambda_{\mathrm{cls}}\mathcal{L}^{\mathrm{cls}} + \mathcal{L}^{\mathrm{aux}}
\end{equation}
with $\mathcal{L}^{\mathrm{mask}}$ being the binary cross-entropy loss, $\mathcal{L}^{\mathrm{dice}}$ being the dice loss, $\mathcal{L}^{\mathrm{cls}}$ being the focal loss, and $\mathcal{L}^{\mathrm{aux}}$ being an auxiliary loss consisting of
\begin{equation}
\mathcal{L}^{\mathrm{aux}} = \lambda_{\mathrm{rank}}\mathcal{L}^{\mathrm{rank}} + \lambda_{\mathrm{seg}}\mathcal{L}^{\mathrm{seg}}
\end{equation}
with $\mathcal{L}^{\mathrm{rank}}$ being the mask-id cross-entropy loss, and $\mathcal{L}^{\mathrm{seg}}$ being the cross-entropy loss.

\subsection{RT-K-Net Architecture}
\label{sec:ShortKNet}

Our RT-K-Net architecture is based on K-Net but introduces several improvements to address shortcomings of K-Net regarding inference speed and accuracy.

\textbf{Mixed Precision Training and Inference:}
One way to speed up the inference of neural networks is to use mixed precision inference.
With mixed precision, operations such as convolutional layers are calculated using float16 instead of float32 precision.
Using mixed precision not only speeds up model inference but also reduces the memory footprint of the model.
We found that using mixed precision with \mbox{K-Net} leads to training divergence.
Since group features are calculated using an element-wise multiplication of features and binary masks, values exceed the maximum value that can be represented using a float16.
We introduce a simple and effective mask normalization strategy to solve this problem in the group feature assembling step in the kernel updates by normalizing the binarized masks $S_{i-1}$ by their mask area
\begin{subequations}
\begin{gather}
	F^{K} = \sum_{u}^{H}\sum_{v}^{W} \frac{S_{i-1}(u, v)}{\sum_{HW}S_{i-1}(u, v)} \cdot F(u, v), \\
    \text{with } S_{i-1}(u, v) = \sigma(M_{i-1}(u, v)) > 0.5.
\end{gather}
\end{subequations}
This normalization not only keeps group feature values within float16 range for mixed precision calculations but also eases training convergence.

\textbf{Post-Processing:}
\begin{algorithm}[tbh]
\caption{Optimized mask pasting strategy}
\label{alg:pasting-strategy}
Generate one hot encoded masks $M_o$ based of $M_{ID}$ \\
Calculate areas in parallel using the one hot encoded masks $M_o$ and the original masks $M$ \\
Filter masks with an overlap below $\delta_o$ \\
Arrange a vector of instance IDs $i \leftarrow [1, 2, ..., N]$ \\
Set instance IDs of stuff classes to zero \\
\tcc{\eg $i \leftarrow [1, 2, 0, 4, 0, 6, ..., N]$}
Generate unique panoptic IDs $l \leftarrow l * \mathrm{offset} + i$ \\
\tcc{$\mathrm{offset}$ is a large constant, \eg 1000}
Generate the final panoptic segmentation $P \leftarrow \sum_{i=0}^{|l|} M_o(i) \cdot l(i)$
\end{algorithm}
K-Net uses a post-processing module similar to previous works~\cite{Cheng2021}, which uses two filtering stages to reduce false positives in the panoptic prediction.
The post-processing consists of the following steps:
\begin{enumerate}
\item 
The class probability prediction~$p\in\mathbb{R}^{N\times N_c}$ is converted into a label prediction~$l$ with corresponding confidence scores~$s$ using a max operation over the class dimension~$N_c$
\begin{equation}
    s = \underset{N_c}{\mathrm{max}}(p),\quad l = \underset{N_c}{\mathrm{argmax}}(p).
\end{equation}
\item
Mask predictions $M\in\mathbb{R}^{N\times H\times W}$ with a confidence score~$s$ below a score threshold $\delta_s$ are filtered out.
\item
Mask IDs are generated by multiplying sigmoid masks with class scores and using the $\mathrm{argmax}$ operation
\begin{equation}
     M_{ID} = \underset{N}{\mathrm{argmax}}(s\cdot\sigma(M)).
\end{equation}
\item
Mask IDs are filtered iteratively by calculating the overlap between the mask area and the original mask area, \ie, $\sigma(M_i) \geq 0.5$.
If the overlap exceeds a threshold $\delta_o$, the mask is pasted to the final panoptic segmentation result~$P$ using the label prediction~$l_i$.
K-Net adds a step by sorting masks based on their confidence scores before iterating over them.
\end{enumerate}
The disadvantage of this post-processing module is that each mask is pasted separately to the final panoptic segmentation result.
Our optimized mask pasting strategy algorithm is shown in Algorithm~\ref{alg:pasting-strategy}.
It alleviates the iterative approach by parallelizing mask filtering and mask pasting.
For mask pasting, a unique label vector is first arranged, then used to generate the final panoptic prediction using element-wise multiplication of $M_o$ with $l$.
Although our post-processing does not support the mask sorting step as done in K-Net, we find that this step does not improve results for RT-K-Net.

\textbf{Backbone and Neck Selection:}
K-Net uses a backbone and neck network for feature extraction.
In K-Net, the authors use ResNet-50-FPN~\cite{Lin2017} as the backbone network and SemanticFPN~\cite{Kirillov2019_1} as the neck for panoptic segmentation.
While this combination is widely used as a baseline, it is not optimized for real-time applications.
Instead, we use the recently introduced RTFormer~\cite{Wang2022} architecture in \mbox{RT-K-Net}.
RTFormer uses an efficient dual-resolution transformer architecture to set a new state-of-the-art in real-time semantic segmentation.
We refer to~\cite{Wang2022} for details on the RTFormer architecture.

\begin{figure}[t!]
\centering
\subfloat[K-Net]{%
    \label{fig:kernel_init:knet}%
\begin{tikzpicture}[node distance=1.2cm, every node/.style={font=\scriptsize}]

\node (convlayer_th) [convlayer, fill=Tan!20] {Conv+BN+ReLU};
\node (convop_th) [conv, right of=convlayer_th, xshift=1cm, fill=Tan!20] {\normalsize $\ast$};
\node (kernels_th) [below of=convop_th, yshift=0.35cm] {$K^{\mathrm{inst}}$};
\node (features_th) [left of=kernels_th, xshift=-2.7cm] {$F^{\mathrm{inst}}$};
\node (masks_th) [right of=kernels_th, xshift=0.5cm] {$M^{\mathrm{inst}}$};

\draw [->,>=stealth] (convlayer_th) -- (convop_th);
\draw [->,>=stealth] (features_th) |- (convlayer_th);
\draw [->,>=stealth] (kernels_th) -- (convop_th);
\draw [->,>=stealth] (convop_th) -| (masks_th);

\node (convlayer_seg) [convlayer, below of=convlayer_th, yshift=-2.2cm, fill=Tan!20] {Conv+BN+ReLU};
\node (convop_seg) [conv, right of=convlayer_seg, xshift=1cm, fill=Tan!20] {\normalsize $\ast$};
\node (kernels_seg) [above of=convop_seg, yshift=-0.35cm] {$K^{\mathrm{seg}}$};
\node (features_seg) [left of=kernels_seg, xshift=-2.7cm] {$F^{\mathrm{seg}}$};
\node (masks_seg) [right of=kernels_seg, xshift=0.5cm] {$M^{\mathrm{seg}}$};
\node (loss_seg) [right of=masks_seg, xshift=0.5cm] {$\mathcal{L}^{\mathrm{seg}}$};

\draw [->,>=stealth] (convlayer_seg) -- (convop_seg);
\draw [->,>=stealth] (features_seg) |- (convlayer_seg);
\draw [->,>=stealth] (kernels_seg) -- (convop_seg);
\draw [->,>=stealth] (convop_seg) -| (masks_seg);
\draw [->,>=stealth,dashed] (loss_seg) -- (masks_seg);

\node (plus) [conv, below of=features_th, yshift=0.35cm, fill=Tan!20] {\tiny $+$};
\node (concat_k) [conv, below of=convop_th, yshift=-0.5cm, fill=Tan!20] {\tiny $\mathbin\Vert$};
\node (concat_m) [conv, below of=masks_th, yshift=0.35cm, fill=Tan!20] {\tiny $\mathbin\Vert$};
\node (f) [right of=plus] {$F$};
\node (k) [right of=f, xshift=2.5cm] {$K_0$};
\node (m) [right of=k, xshift=0.5cm] {$M_0$};

\draw [->,>=stealth] (features_th) -- (plus);
\draw [->,>=stealth] (features_seg) -- (plus);
\draw [->,>=stealth] (plus) -- (f);

\draw [->,>=stealth] (kernels_th) -- (concat_k);
\draw [->,>=stealth] (kernels_seg) -- (concat_k);
\draw [->,>=stealth] (concat_k) -- (k);

\draw [->,>=stealth] (masks_th) -- (concat_m);
\draw [->,>=stealth] (masks_seg) -- (concat_m);
\draw [->,>=stealth] (concat_m) -- (m);

\end{tikzpicture}%
}
\hfill
\subfloat[RT-K-Net]{%
    \label{fig:kernel_init:rtknet}%
\begin{tikzpicture}[node distance=1.2cm, every node/.style={font=\scriptsize}]

\node (convlayer_pan) [convlayer, fill=Tan!20] {Conv+BN+ReLU};
\node (convop_pan) [conv, right of=convlayer_pan, xshift=1cm, fill=Tan!20] {\normalsize $\ast$};
\node (kernels_pan) [below of=convop_pan, yshift=0.35cm] {$K_0$};
\node (features_pan) [left of=kernels_pan, xshift=-2.7cm] {$F$};
\node (masks_pan) [right of=kernels_pan, xshift=0.5cm] {$M_0$};

\draw [->,>=stealth] (convlayer_pan) -- (convop_pan);
\draw [->,>=stealth] (features_pan) |- (convlayer_pan);
\draw [->,>=stealth] (kernels_pan) -- (convop_pan);
\draw [->,>=stealth] (convop_pan) -| (masks_pan);

\node (aux_head) [aux_head, below of=convlayer_pan, yshift=-0.8cm, xshift=1.5cm] {};
\node[anchor=south east,inner sep=2pt] at (aux_head.south east) {Aux Seg Head};
\node (seg) [convlayer, below of=convlayer_pan, yshift=-0.8cm] {Conv+BN+ReLU};
\node (seg_conv) [conv, right of=seg, xshift=1cm] {\normalsize $\ast$};
\node (seg_mask) [right of=seg_conv, xshift=0.5cm] {$M^{\mathrm{seg}}$};+
\node (loss_mask) [right of=seg_mask, xshift=0.5cm] {$\mathcal{L}^{\mathrm{seg}}$};

\draw [->,>=stealth] (features_pan) |- (seg);
\draw [->,>=stealth] (seg) -- (seg_conv);
\draw [->,>=stealth] (seg_conv) -- (seg_mask);
\draw [->,>=stealth,dashed] (loss_mask) -- (seg_mask);

\end{tikzpicture}%
}
\caption{
Kernel and mask initialization used in (a) K-Net and (b) RT-K-Net.
Convolution operations are visualized as $\ast$, concatenations as $\mathbin\Vert$.
RT-K-Net uses a much simpler approach without separate branches for semantic and instance kernels than K-Net. 
Instead, RT-K-Net uses an auxiliary semantic head, which can be omitted during inference.
}
\label{fig:kernel_init}
\end{figure}
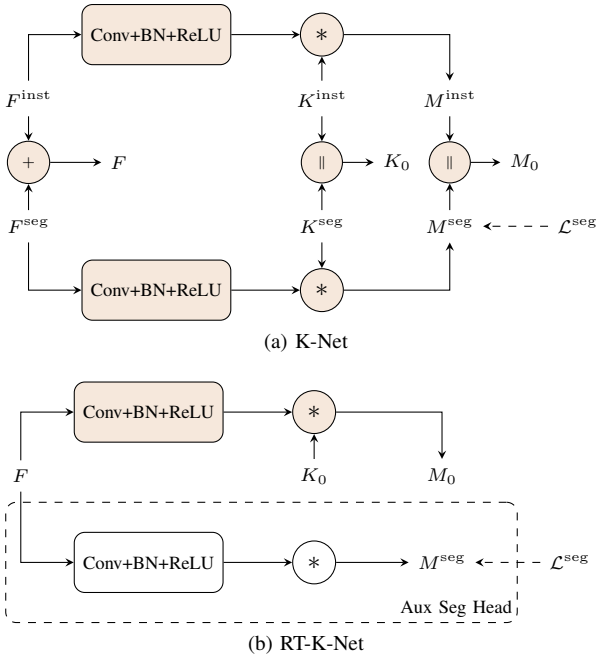

\textbf{Mask and Kernel Initialization:}
K-Net uses two branches for mask and kernel initialization, one for semantic segmentation and one for instance segmentation.
The initialization of K-Net is shown in Fig.~\ref{fig:kernel_init:knet}.
Each branch uses a feature map $F^{\mathrm{inst}}$ or rather $F^{\mathrm{seg}}$, initial kernels $K^{\mathrm{inst}}$ and $K^{\mathrm{seg}}$, and initial mask predictions $M^{\mathrm{inst}}$ and $M^{\mathrm{seg}}$.
The mask predictions $M^{\mathrm{inst}}$ and $M^{\mathrm{seg}}$ are generated by performing convolution of $F^{\mathrm{inst}}$ with $K^{\mathrm{inst}}$ and $F^{\mathrm{seg}}$ with $K^{\mathrm{seg}}$, respectively.
The feature maps are combined using summation, while kernels and mask predictions are concatenated.
Note that the authors do not split the branches into thing and stuff classes but rather split between instance and semantic segmentation.
This way, they can add an auxiliary loss term~$\mathcal{L}^{\mathrm{seg}}$ for semantic segmentation on $M^{\mathrm{seg}}$.
The authors state that the split into two different branches is unnecessary but yields about $1~\%$ PQ improvement.
We argue that it adds unnecessary complexity to the model and deviates from the idea of a unified architecture.
Hence, we propose a simplified version using only one unified branch.
We found that merging the branches is non-trivial since the auxiliary semantic loss is significant for good performance.
Our proposed initialization method of RT-K-Net is shown in Fig.~\ref{fig:kernel_init:rtknet}.
We initialize panoptic kernels $K_0$ directly and use a single feature map $F$ which is produced by the RTFormer segmentation head.
$M_0$ is generated by performing convolution of the feature map~$F$ with the kernels~$K_0$.
To still be able to add the auxiliary semantic loss, we add an auxiliary semantic segmentation head, which follows the same structure as the main initialization head.
This head is only necessary during training and can be omitted during inference.

\subsection{Improvements for Instance Segmentation Quality}
\label{sec:InstanceTraining}

After describing our RT-K-Net, we introduce two methods to increase the segmentation quality for thing masks.

\begin{figure}[t!]
\begin{center}
    \includegraphics[width=\columnwidth]{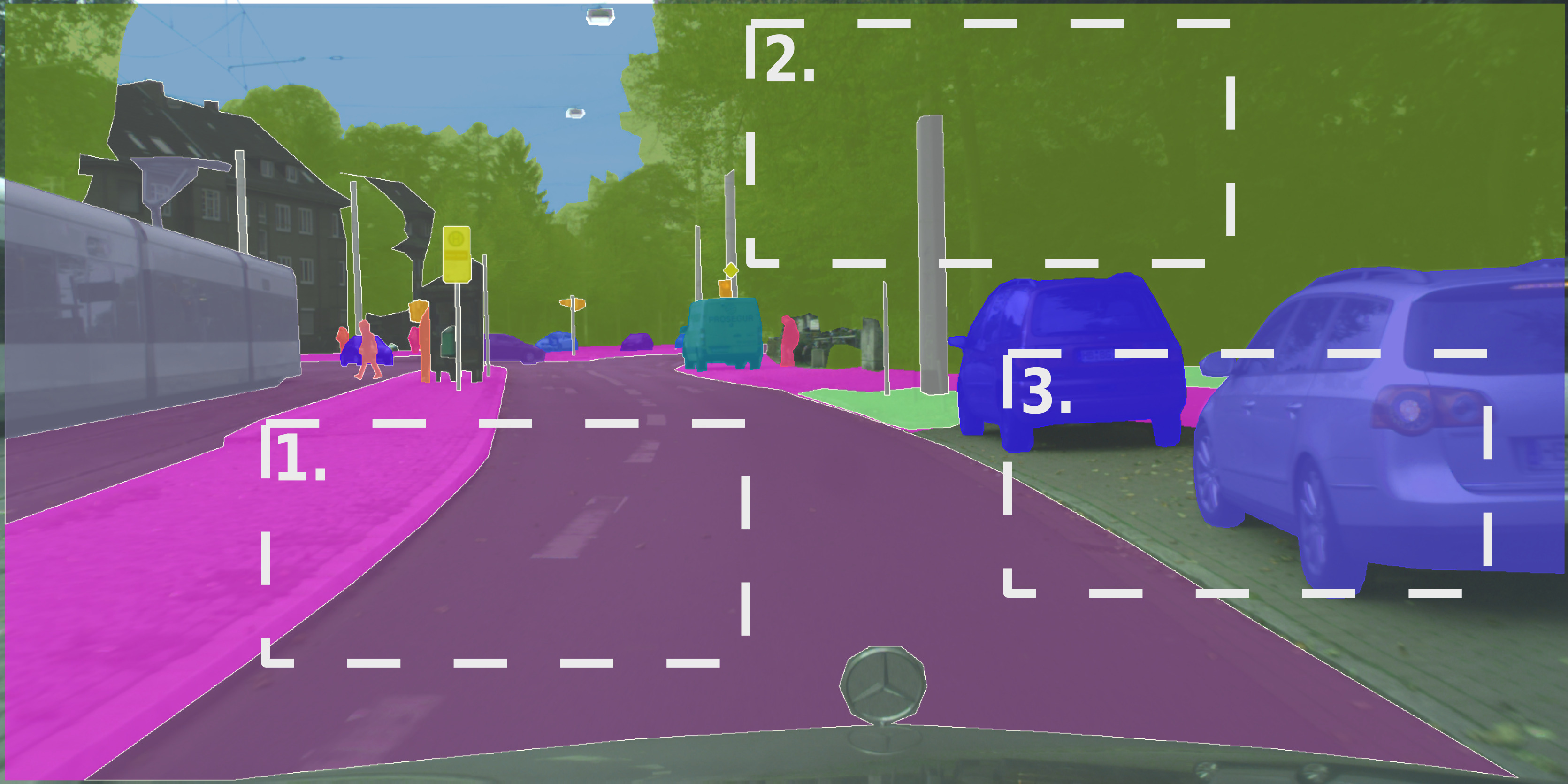}
\end{center}
\vspace{-3mm}
   \caption{Visualization of our instance-aware crop augmentation. The crop window is selected so that, if possible, at least the geometrical center of one thing class mask is inside the crop window. In this example, the third window is selected since it contains the geometrical centers of the two car object masks.}
\label{fig:inst_crop}
\vspace{-3mm}
\end{figure}

\textbf{Instance-aware crop augmentation:}
Crop-based training is the standard for training panoptic segmentation models on high-resolution images.
For crop-based training, images are scaled up or down randomly.
Then, random cropping is used to generate images of the same size for training.
We find that random cropping has the disadvantage that it often crops out all thing masks from the image.
Therefore, during training, RT-K-Net sees fewer examples with thing masks, resulting in a lower panoptic quality for thing classes.
Hence, we develop an instance-aware cropping method.
Instead of taking the first random crop, we check if the crop area contains at least the geometrical center of one thing mask.
If the crop area contains no such center, we iteratively sample crop areas until we find a crop area containing such a center.
Since there are cases where the entire image only contains stuff masks, we abort the algorithm after ten iterations and use a random crop window in that case.
Fig.~\ref{fig:inst_crop} shows an example of our instance-aware crop augmentation.
In this case, the third crop area is used since it contains the centers of both car objects.

\textbf{Instance discrimination loss:}
An accurate mask prediction not only requires discriminative kernels $K$ but also distinct features from the feature map $F$ since kernels are convolved with the feature map to generate the mask prediction.
Following recent methods~\cite{Yu2022, Yu2022_1}, we adopt an auxiliary instance discrimination loss to improve the discrimination ability of the feature map $F$.
Here, we follow CMT-DeepLab~\cite{Yu2022_1} and apply a contrastive loss
\begin{equation}
\label{eq:pixel_insdis_loss}
\mathcal{L}^{\mathrm{inst}}=\sum_{a\in A}\frac{-1}{|P(a)|}\sum_{p\in P(a)} \log \frac{\exp{\left(F_{a}\cdot F_{p} / \tau\right)}}{\sum_{b\in A} \exp{\left(F_{a}\cdot F_{b} / \tau\right)}},
\end{equation}
where $A$ is a sampled set of pixels from the image, $P(a)$ is the subset of $A$ that belongs to the same mask as $a$, $|P(a)|$ is its cardinally, and $\tau$ is the temperature.
The instance discrimination loss is added to the auxiliary loss
\begin{equation}
\mathcal{L}^{\mathrm{aux}} = \lambda_{\mathrm{rank}}\mathcal{L}^{\mathrm{rank}} + \lambda_{\mathrm{seg}}\mathcal{L}^{\mathrm{seg}} + \lambda_{\mathrm{inst}}\mathcal{L}^{\mathrm{inst}}.
\end{equation}
\section{Experimental Results}
In this section, we evaluate RT-K-Net quantitatively and qualitatively.
We provide comparisons with state-of-the-art panoptic segmentation methods and ablation studies of our approach.

\subsection{Datasets}
We conduct experiments on the Cityscapes~\cite{Cordts2016} and the Mapillary Vistas~\cite{Neuhold2017} dataset.
The Cityscapes dataset provides 5000 annotated high resolution $1024 \times 2048$ pixel images with fine-grained panoptic labels for 30 classes, of which 19 (8 thing, 11 stuff) are used during evaluation.
Of the 5000 images, 2975 are used for training, 500 for evaluation, and 1525 for testing.
The Mapillary Vistas dataset provides 25000 high-resolution images with an average of $8.9\times10^6$ pixels per image.
Of the 25000 images, 18000 are used for training, 2000 for evaluation, and 5000 for testing.
We train on version 1.2 of the dataset, consisting of 65 classes (37 thing, 28 stuff).

\subsection{Implementation Details}
 \label{sec:impl}
We implement our method in Pytorch~\cite{Paszke2019} using the detectron2 framework~\cite{Wu2019}.
On Cityscapes, we train on 4 RTX 2080Ti GPUs, while on Mapillary Vistas, we train on 8 RTX 2080Ti GPUs.
We use the AdamW optimizer with the ``poly'' learning rate scheduler~\cite{Liu2015} with a base learning rate of 0.002.
We add warmup, where we linearly increase the learning rate from 0.00002 to 0.002 in the first 1000 iterations.
We use a total batch size of 32, weight decay of 0.05, and gradient clipping with a clip value of 1.0.
We perform ablation studies on Cityscapes, where we train for 60k iterations unless stated otherwise.
For our final results, we train on Cityscapes for 90k iterations and on Mapillary Vistas for 300k iterations.
We use mixed precision training and inference as described in Section~\ref{sec:ShortKNet} and set the number of predicted masks to $N=100$.
For inference, we filter out masks with confidences below $\delta_s=0.3$ and an overlap below $\delta_o=0.6$.
On Mapillary Vistas, we set $\delta_s=0$ to deal with the high number of objects per image.
We do not filter out small stuff classes based on their area or use any test-time augmentations to improve accuracy.
For data augmentation, we use random image scaling between 0.5 and 2.1, random cropping with our instance crop method as described in Section~\ref{sec:InstanceTraining}, random image color and brightness adjustment, and random left-right flip.
For Cityscapes, we use a crop size of $512 \times 1024$ pixel, for Mapillary Vistas, we use a crop size of $1024 \times 1024$ pixel.
For loss weights, we follow the original K-Net settings and set $\lambda_{mask}=1.0, \lambda_{dice}=4.0, \lambda_{rank}=0.1, \lambda_{cls}=2.0$, and $\lambda_{seg}=1.0$.
We use $\lambda_{inst}=1.0$ for the instance discrimination loss.
For $\mathcal{L}^{seg}$, instead of using the cross-entropy loss as used in K-Net, we use the weighted bootstrapped cross entropy loss~\cite{Yang2019} instead.
In contrast to K-Net, we increase the number of kernel updates from 3 to 4 since we found that it does not add much runtime while slightly improving performance.
For RTFormer, we use the Base variant in all our experiments and the ImageNet~\cite{Deng2009} pre-trained weights as initialization.

\subsection{Ablation Studies}
\begin{table}
\caption{Ablation studies on Cityscapes.}
\label{tab:ablation}
\begin{center}
\begin{tabular}{lcccccccc}
\toprule
Method & PQ $\uparrow$ & PQ$_{\text{th}}$ $\uparrow$ & PQ$_{\text{st}}$ $\uparrow$ & Runtime in ms $\downarrow$ \\
\midrule
K-Net~\cite{Zhang2021} & 56.9 & 46.0 & 64.8 & 112 \\
+ Mask Norm. (FP16) & 57.8 & 48.7 & 64.4 & 66 \\
+ Opt Post-Proc & 57.8 & 48.7 & 64.4 & 53 \\
+ RTFormer & 58.5 & 47.3 & 66.6 & 35 \\
+ Merged Kernels & 57.7 & 47.6 & 65.0 & 32 \\
+ Aux Seg Head & 58.9 & 48.9 & 66.2 & 32 \\
+ Inst Crop Aug & 59.3 & 49.7 & 66.4 & 32 \\
+ Inst Disc Loss & 59.8 & 51.3 & 66.0 & 32 \\
+ 90k Schedule & \textbf{60.2} & \textbf{51.5} & \textbf{66.5} & \textbf{32} \\
\bottomrule
\end{tabular}
\end{center}
\vspace{-3mm}
\end{table}
We perform ablation studies on the Cityscapes dataset.
We start by training the original K-Net on Cityscapes using our training configuration described in Section~\ref{sec:impl}.
K-Net reaches a PQ of $56.9\%$ and a runtime of $112~\si{ms}$.
Adding our mask normalization as described in Section~\ref{sec:ShortKNet} enables us to use mixed precision training and inference.
Therefore, the runtime drops from $112~\si{ms}$ to $66~\si{ms}$.
In addition, we find that our mask normalization improves panoptic quality by $0.9\%$ PQ.
Next, we add our optimized post-processing method, which decreases the runtime to $53~\si{ms}$ without sacrificing performance.
Replacing the backbone and the neck with \mbox{RTFormer} boosts performance from $57.8\%$ to $58.5\%$ PQ while reducing runtime to $35~\si{ms}$.
Next, we replace the mask and kernel initialization with our simplified version.
Panoptic quality drops from $58.5\%$ to $57.7\%$ PQ when using our simplified version without adding the auxiliary segmentation head but increases to $58.9\%$ PQ when adding the auxiliary segmentation head.
Adding our instance crop augmentation and instance discrimination loss, we can boost the panoptic quality of thing classes from $48.9\%$ to $51.3\%$ PQ$_{\text{th}}$.
Finally, by extending the training schedule to 90k iterations, we achieve our final panoptic result of $60.2\%$ PQ.

\subsection{Main Results}
\begin{figure*}[t]
\centering
\subfloat{%
    \includegraphics[width=0.24\linewidth]{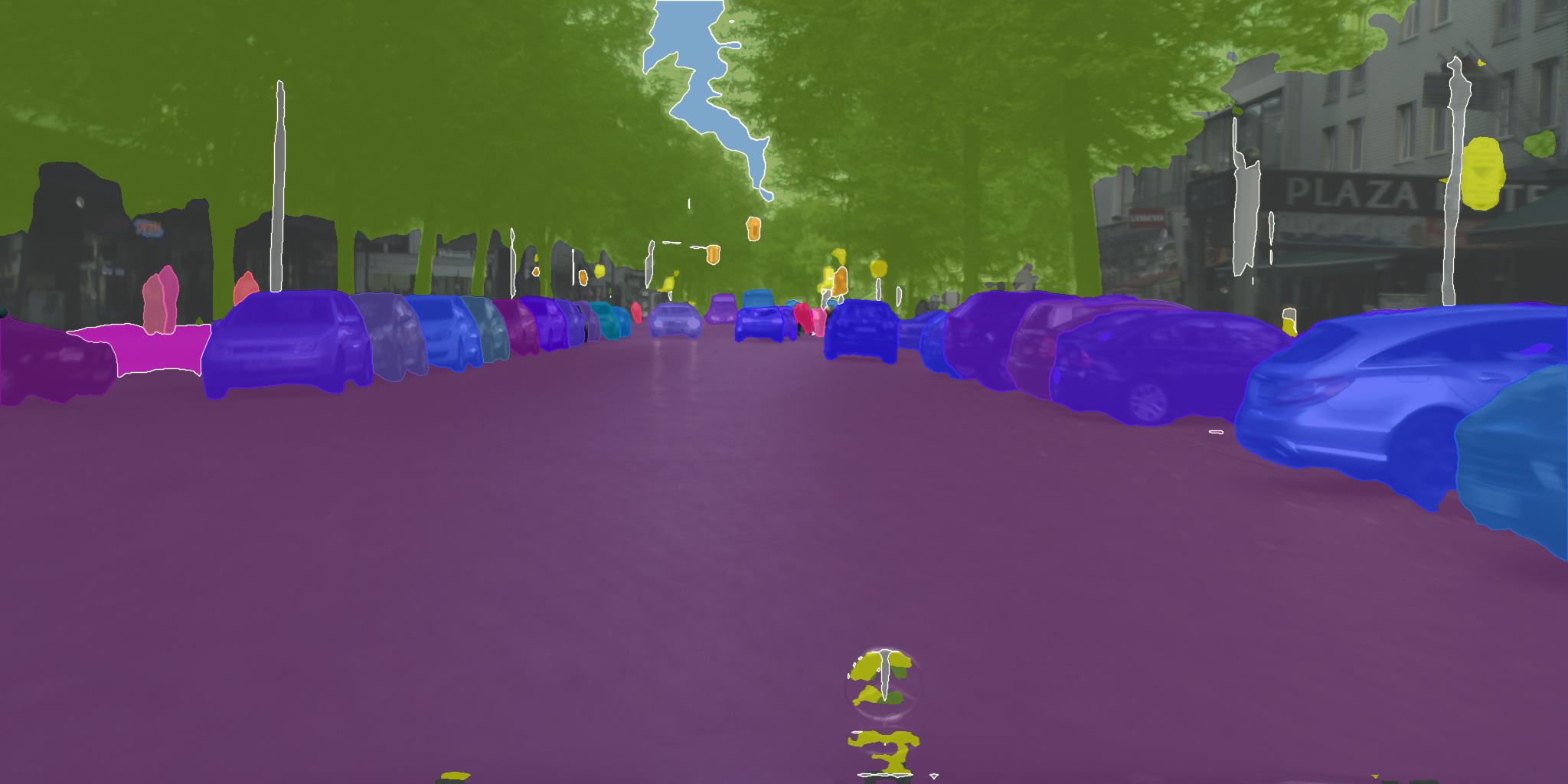}%
}
\hfill
\subfloat{%
    \includegraphics[width=0.24\linewidth]{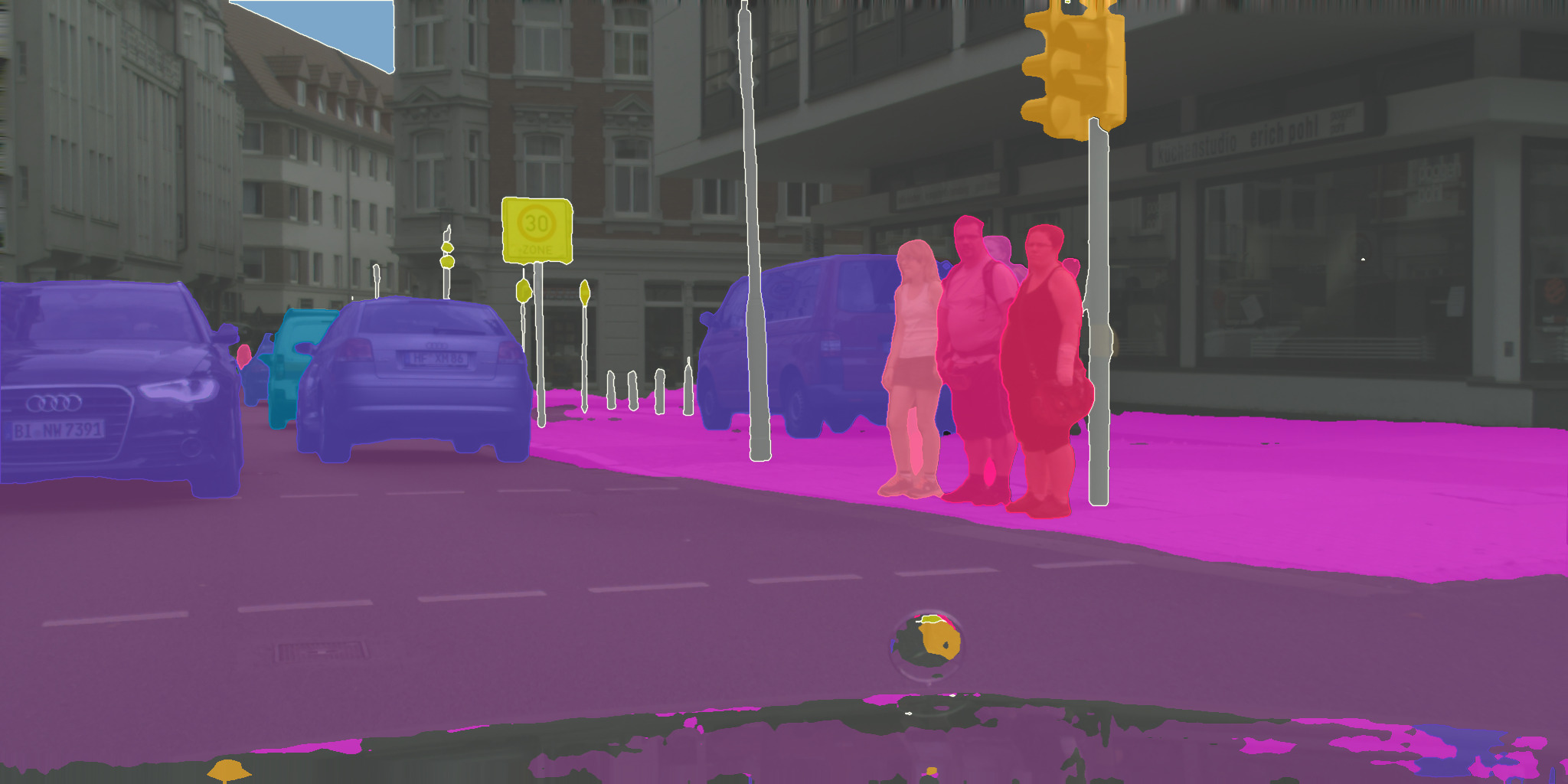}%
}
\hfill
\subfloat{%
    \includegraphics[width=0.24\linewidth]{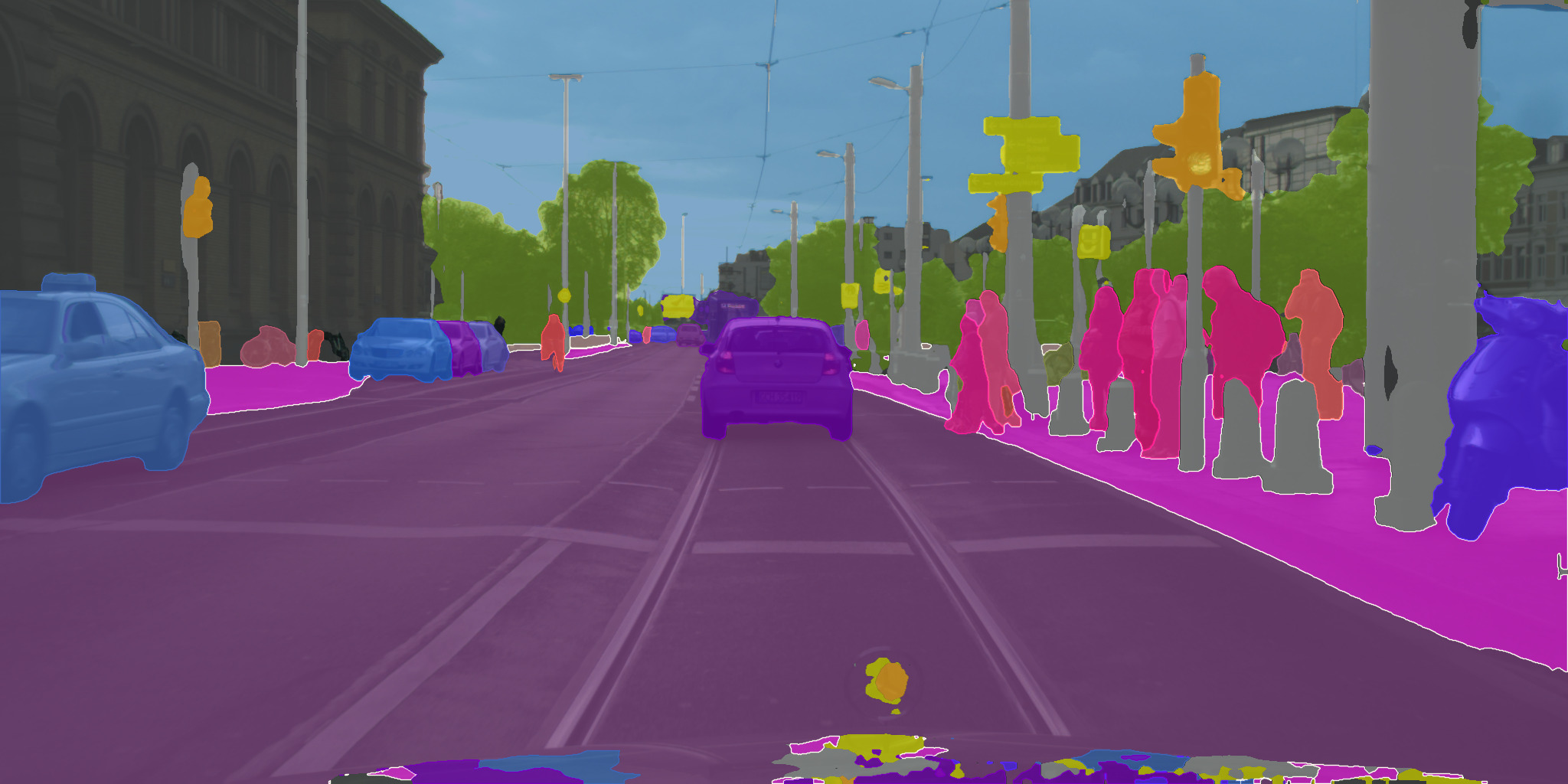}%
}
\hfill
\subfloat{%
    \includegraphics[width=0.24\linewidth]{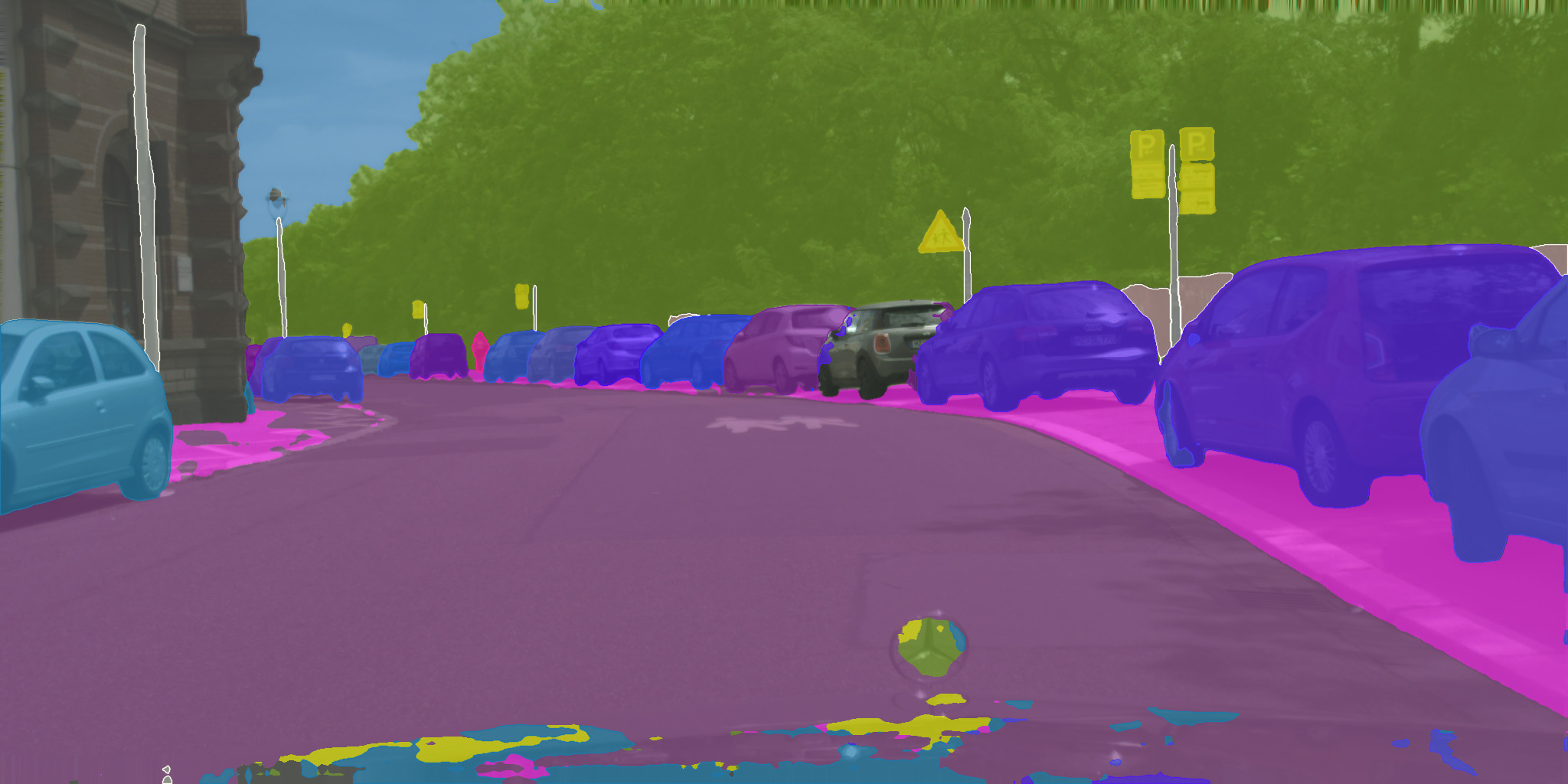}%
}
\vfill
\subfloat{%
    \includegraphics[width=0.24\linewidth]{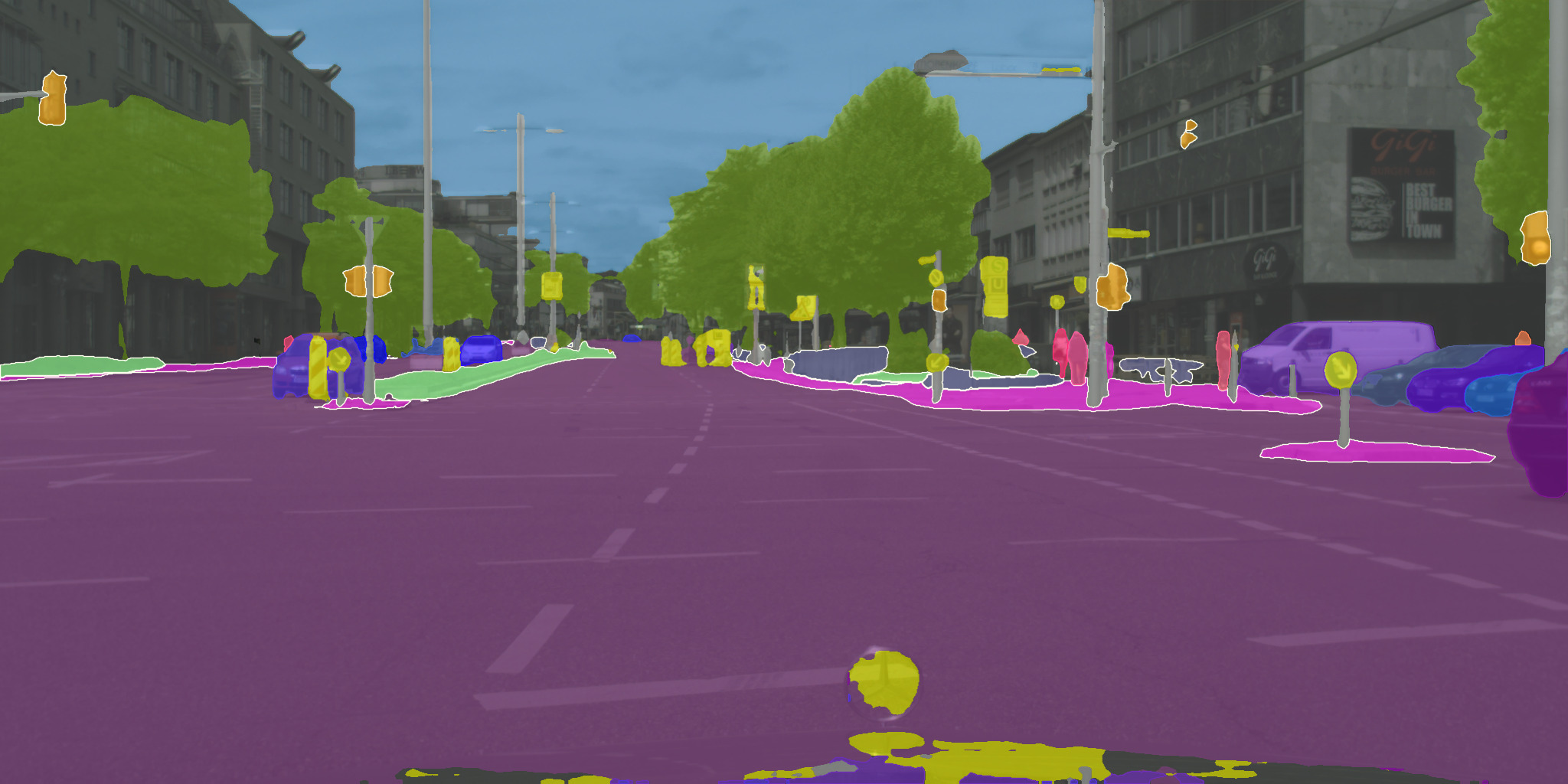}%
}
\hfill
\subfloat{%
    \includegraphics[width=0.24\linewidth]{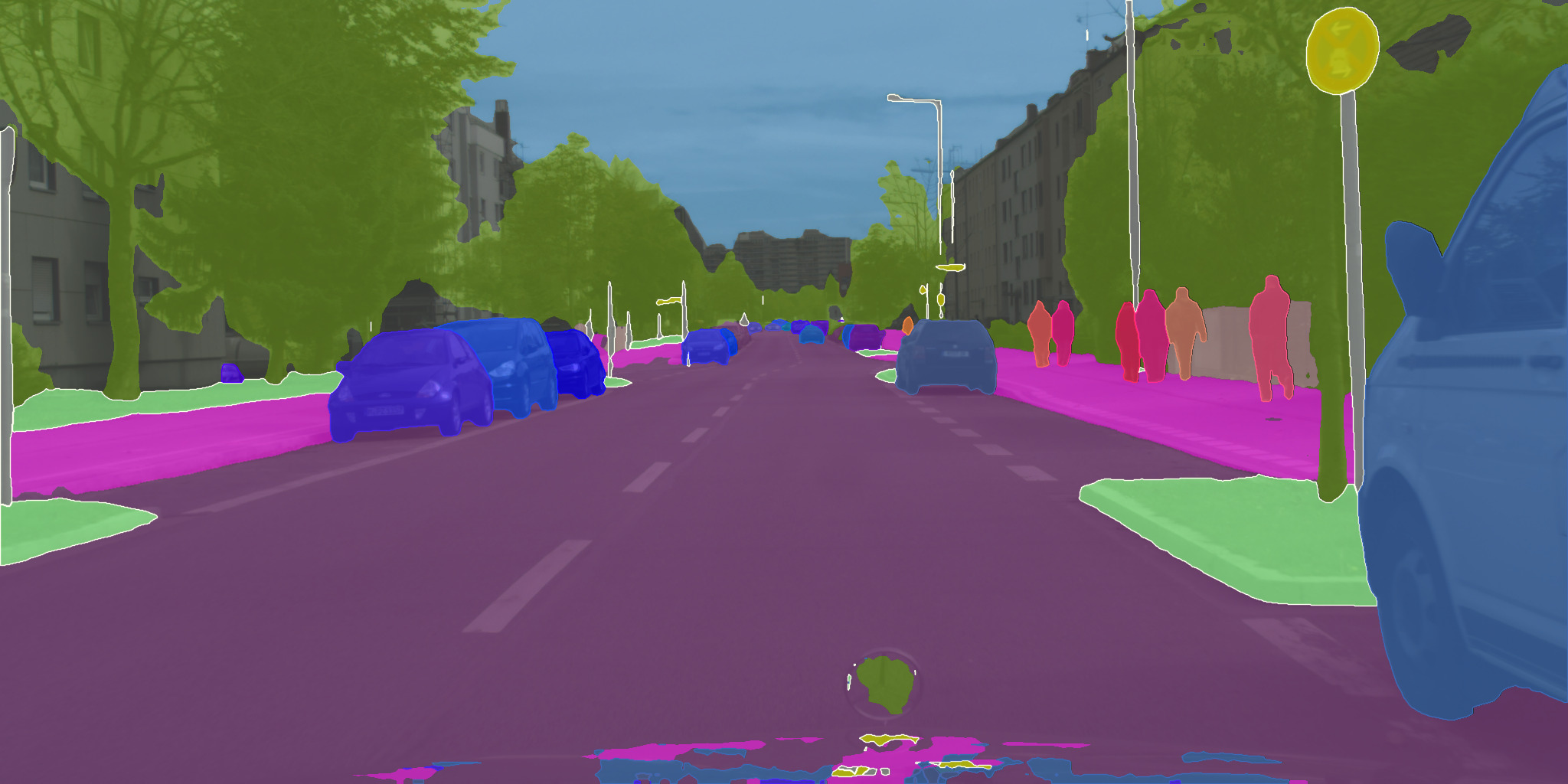}%
}
\hfill
\subfloat{%
    \includegraphics[width=0.24\linewidth]{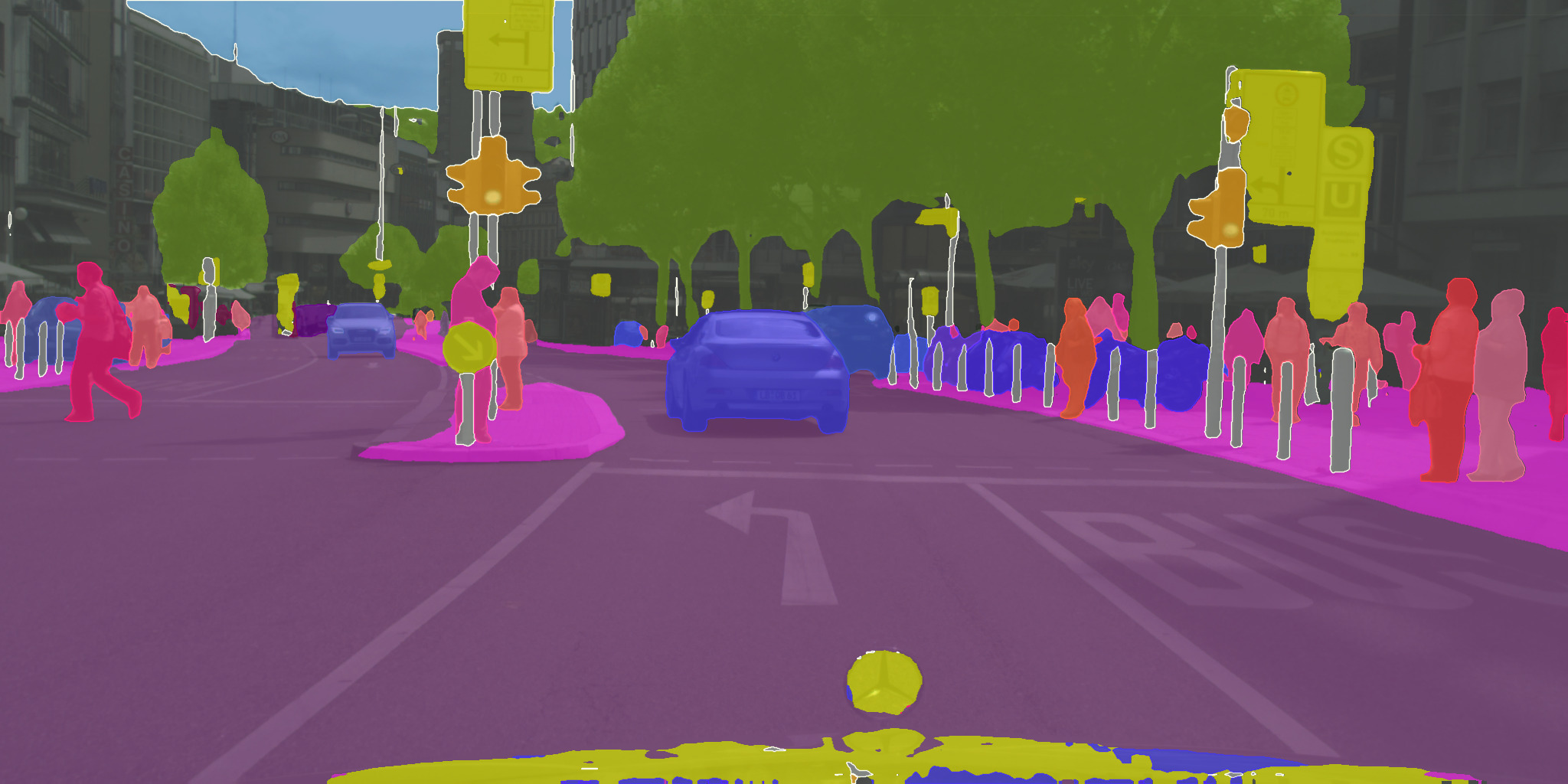}%
}
\hfill
\subfloat{%
    \includegraphics[width=0.24\linewidth]{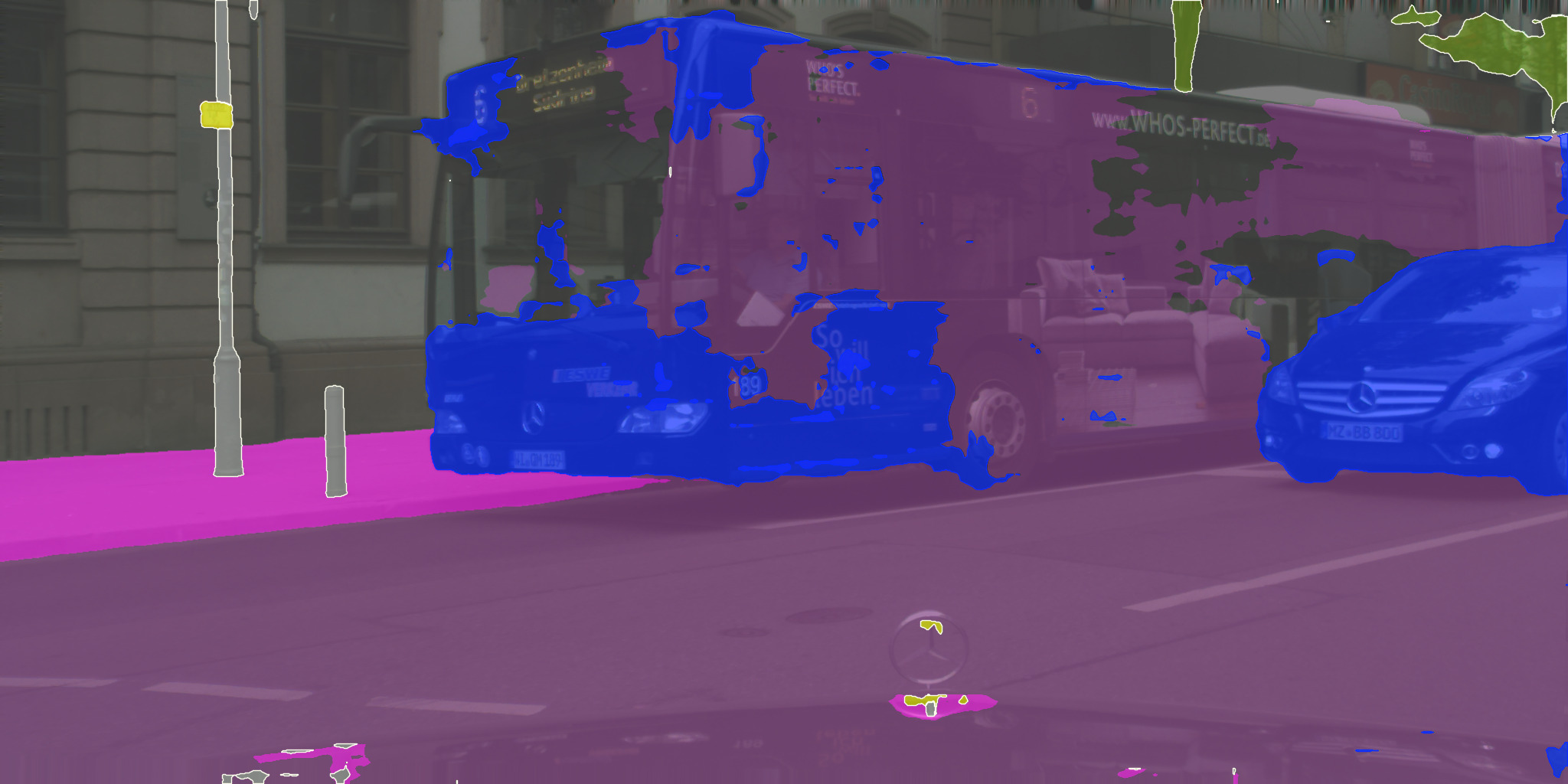}%
}
\vfill
\subfloat{%
    \includegraphics[width=0.24\linewidth]{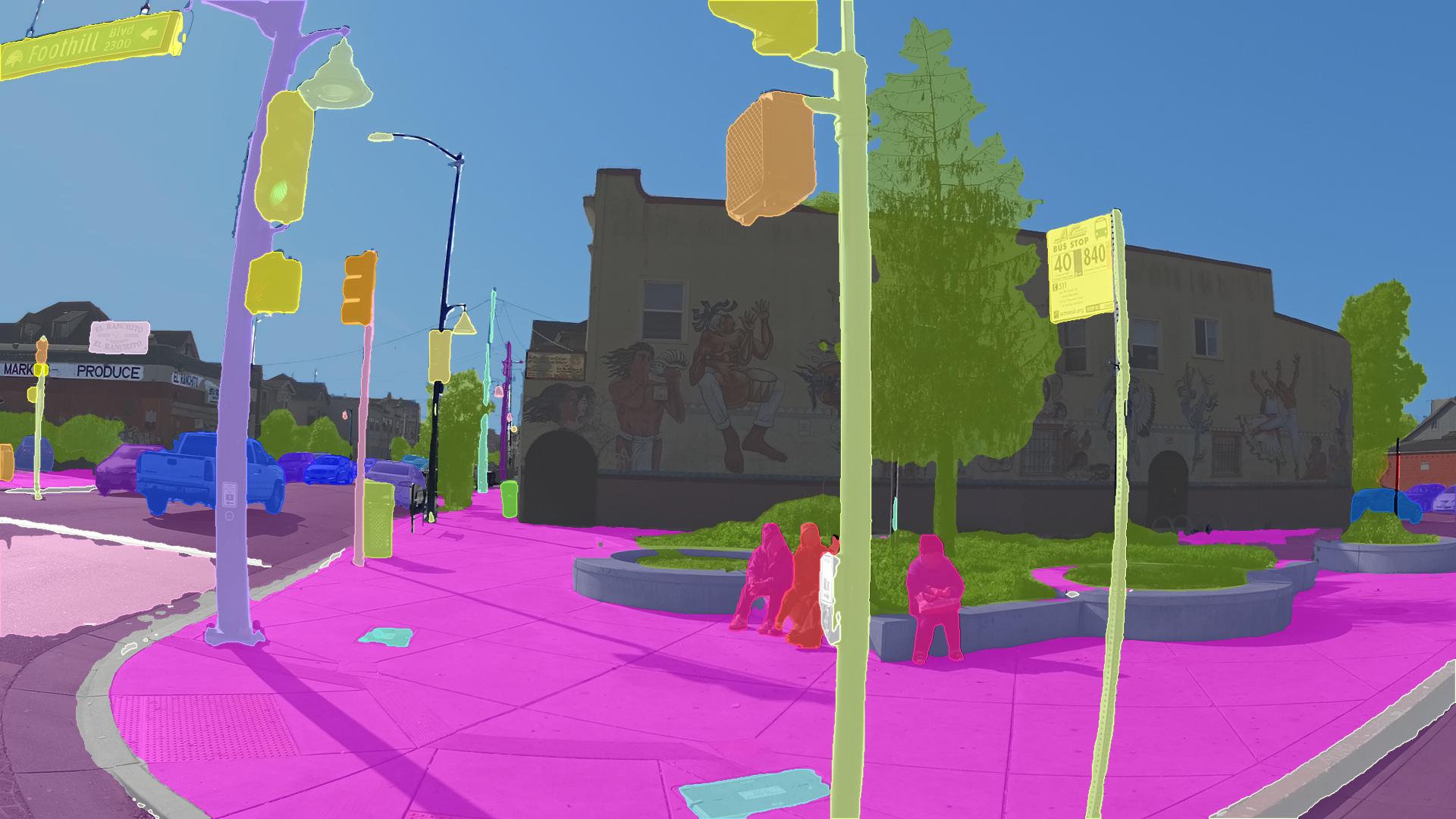}%
}
\hfill
\subfloat{%
    \includegraphics[width=0.24\linewidth]{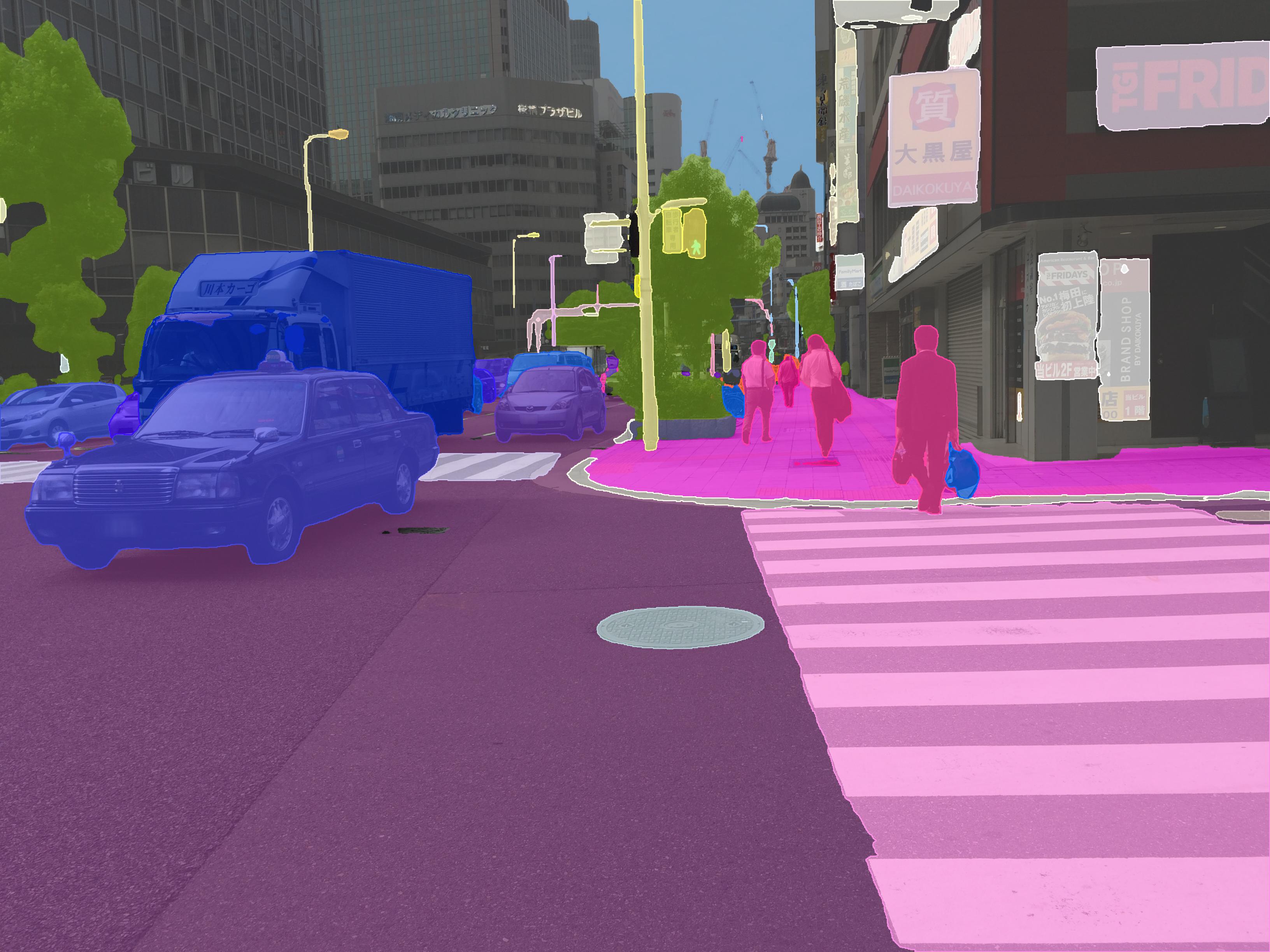}%
}
\hfill
\subfloat{%
    \includegraphics[width=0.24\linewidth]{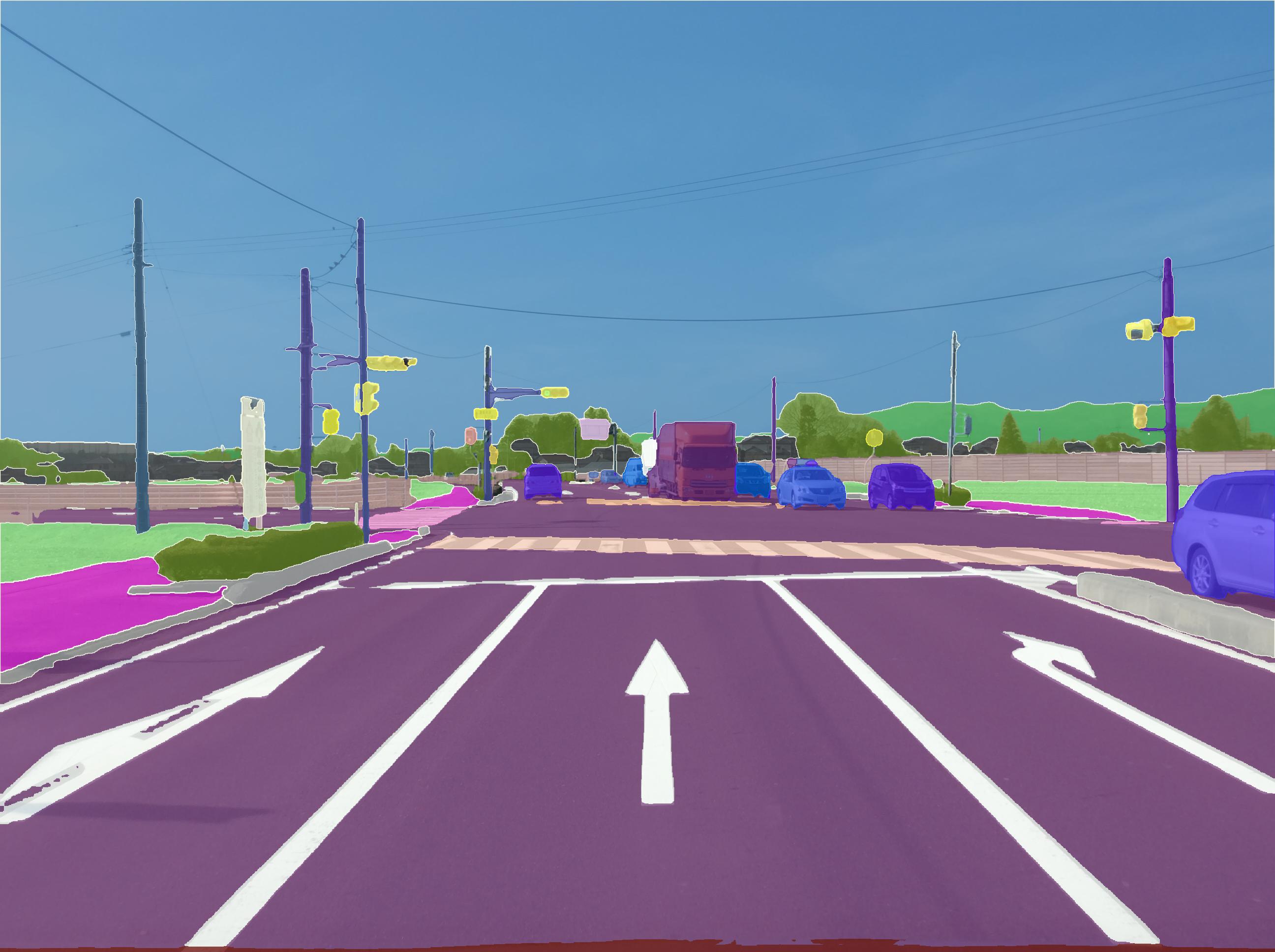}%
}
\hfill
\subfloat{%
    \includegraphics[width=0.24\linewidth]{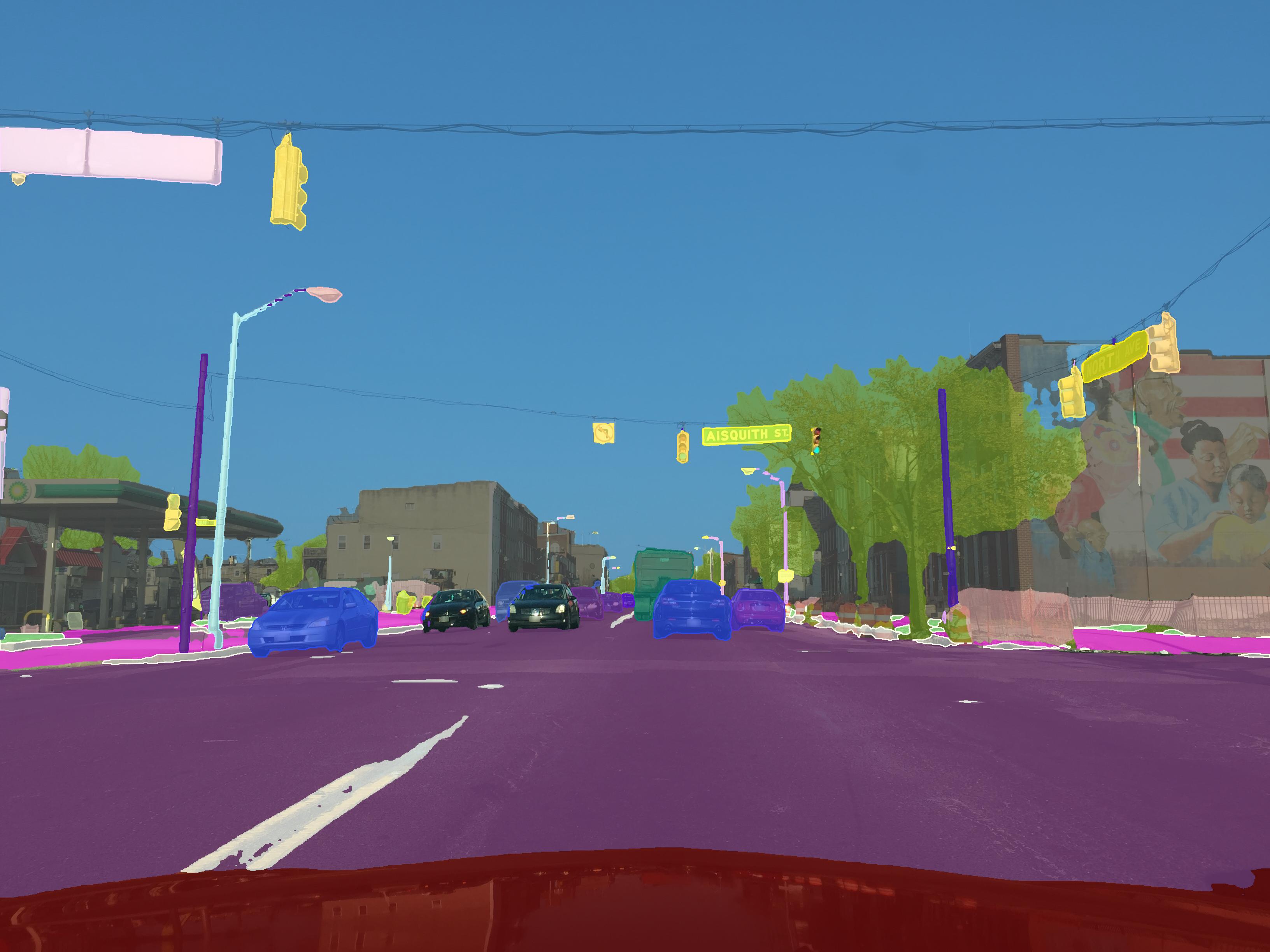}%
}
\vfill
\subfloat{%
    \includegraphics[width=0.24\linewidth]{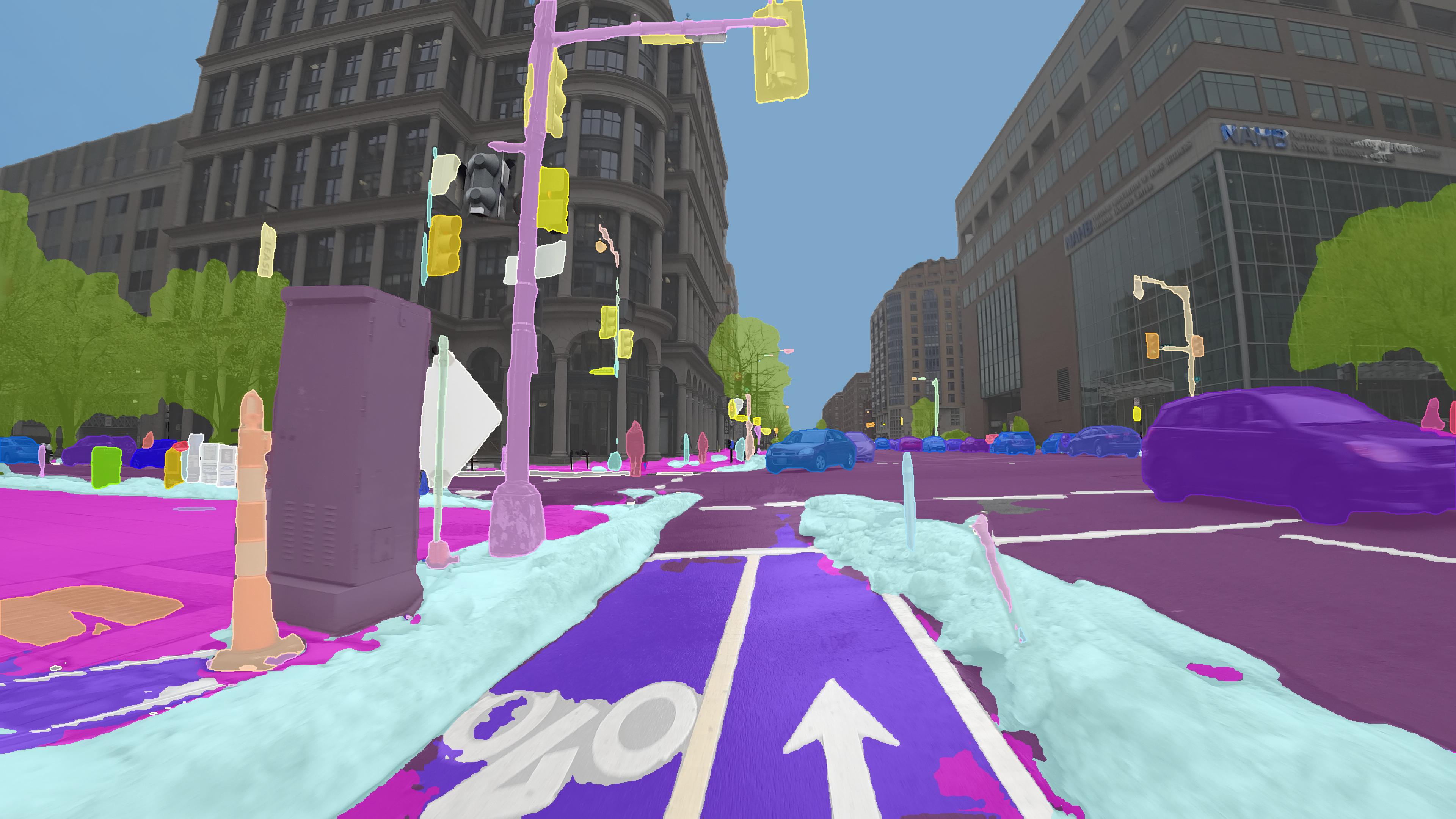}%
}
\hfill
\subfloat{%
    \includegraphics[width=0.24\linewidth]{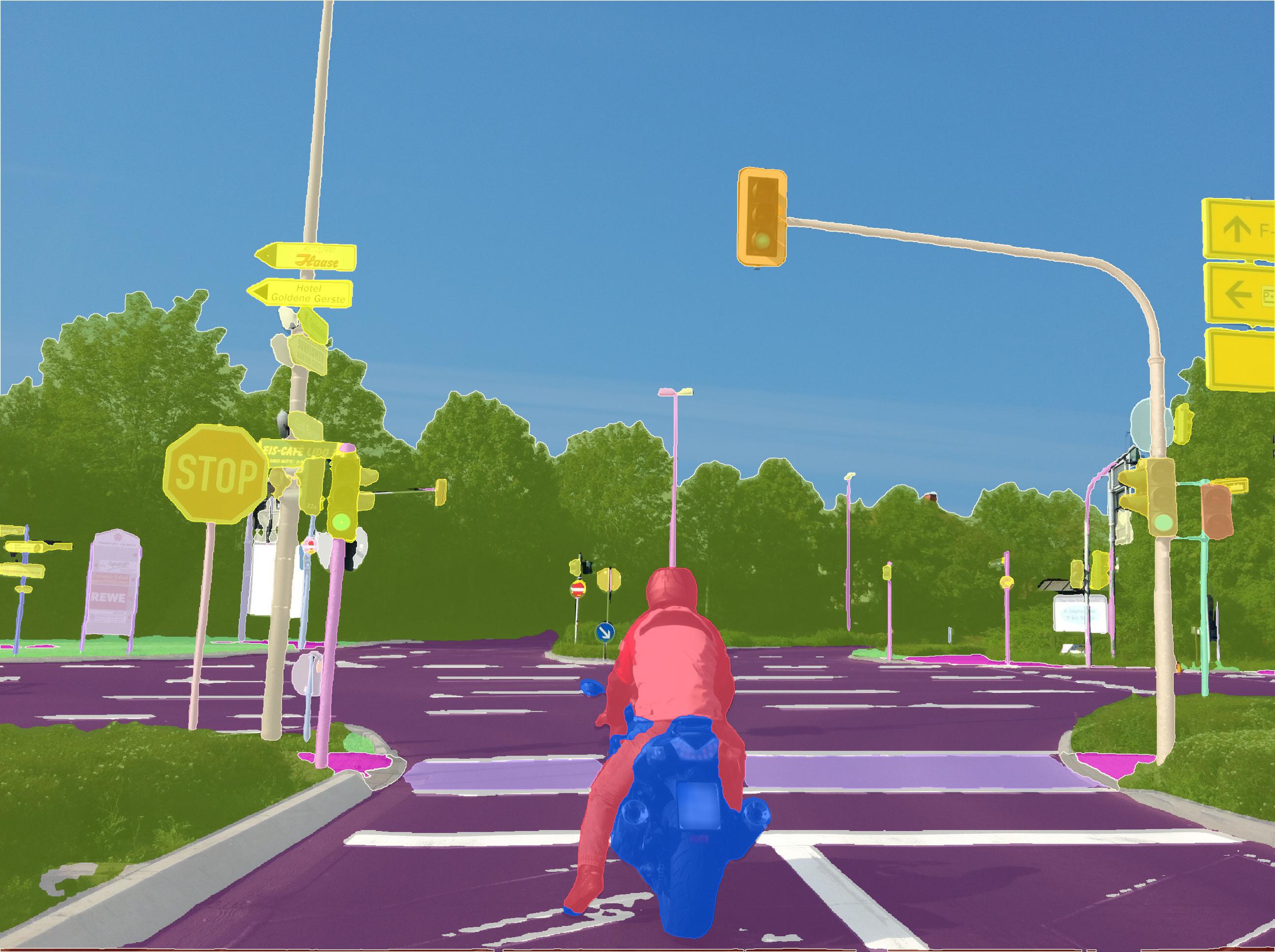}%
}
\hfill
\subfloat{%
    \includegraphics[width=0.24\linewidth]{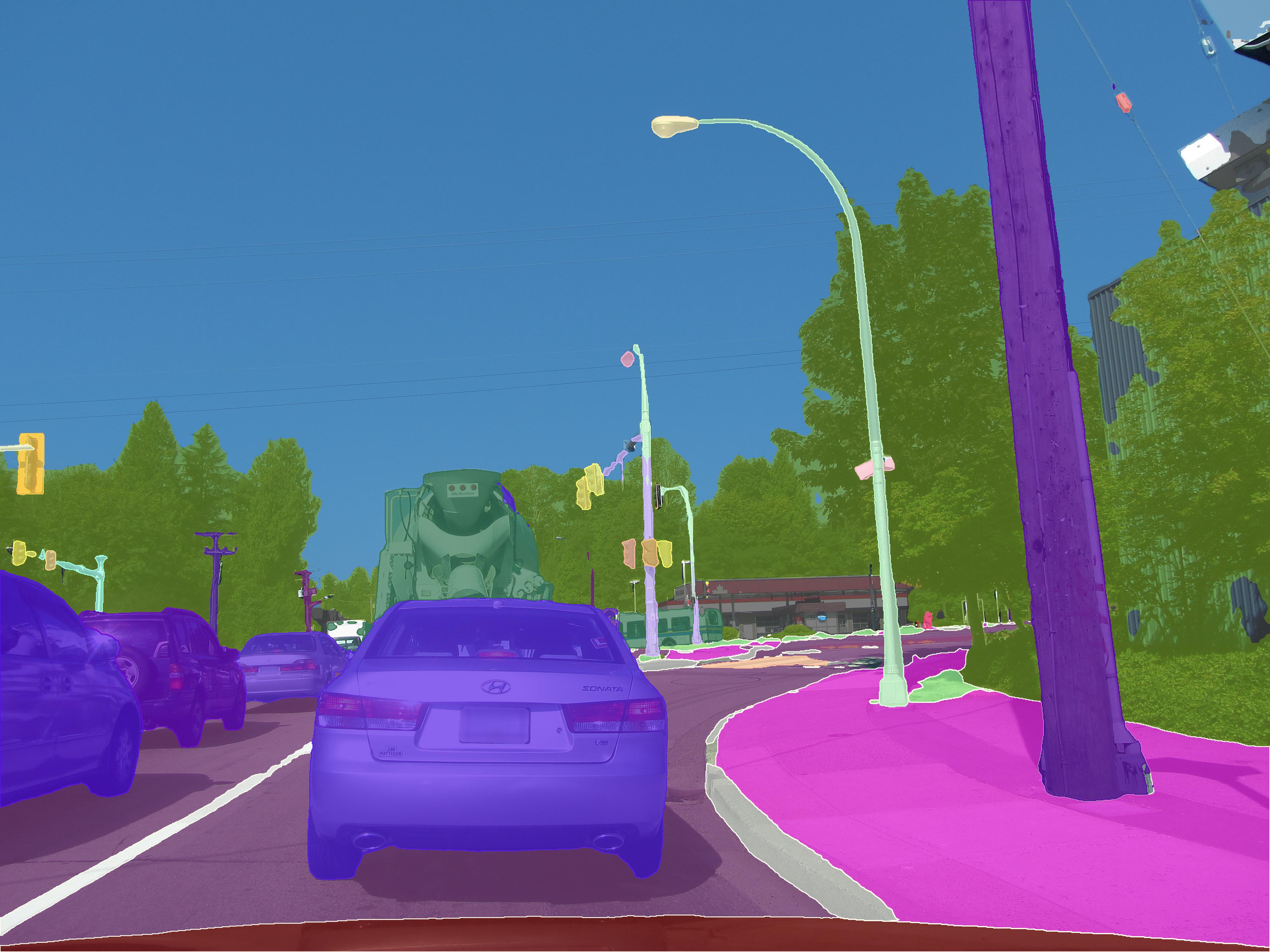}%
}
\hfill
\subfloat{%
    \includegraphics[width=0.24\linewidth]{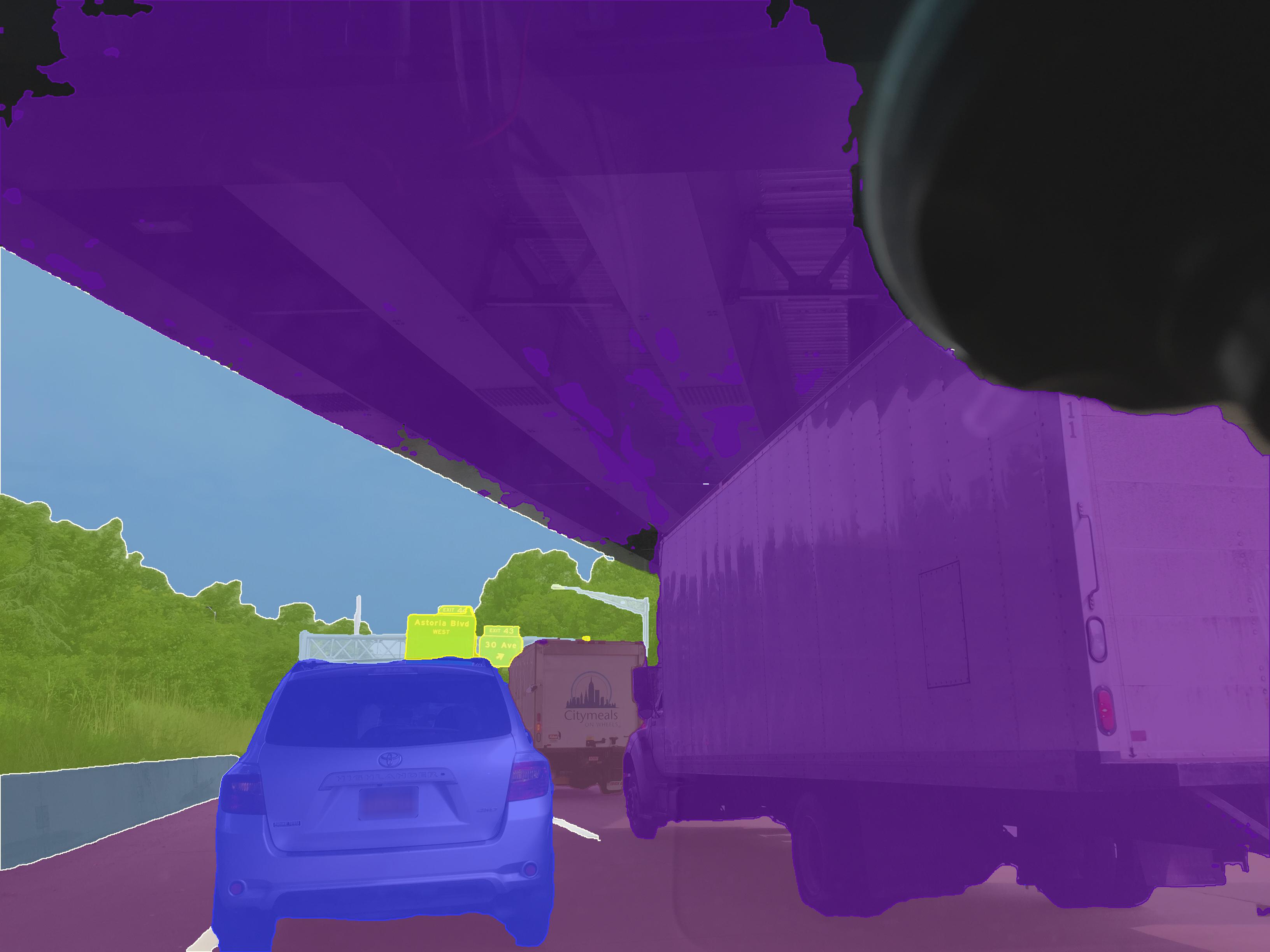}%
}
\caption{Qualitative results on unseen images from the Cityscapes dataset~\cite{Cordts2016} (first two rows) and the Mapillary Vistas dataset~\cite{Neuhold2017} (last two rows). Each mask prediction gets a unique color. The last column shows misclassifications, where RT-K-Net has problems finding all objects or predicting an accurate mask for rare classes.}
\label{fig:qualitative}
\end{figure*}

\begin{table*}
\caption{Comparison of our method to state-of-the-art real-time panoptic segmentation methods on the Cityscapes validation set. Best results are shown in \textbf{boldface}, second best \underline{underlined}.}
\label{tab:cityscapes}
\begin{center}
\begin{tabular}{lcccccc}
\toprule
Method & Backbone & PQ $\uparrow$ & PQ$_{\text{th}}$ $\uparrow$ & PQ$_{\text{st}}$ $\uparrow$ & GPU &  Runtime in ms $\downarrow$ \\
\midrule
MGNet~\cite{Schoen2021} & ResNet18 & 55.7 & 45.3 & 63.1 & Titan RTX & \underline{36} \\
Petrovai and Nedevschi~\cite{Petrovai2020} & VoVNet2-39 & 57.3 & 50.4 & 62.4 & V100 & 82 \\
Panoptic-DeepLab~\cite{Chen2020} & SWideRNet-(0.25, 0.25, 0.75) & 58.4 & - & - & V100 & 63 \\
Hou~\etal~\cite{Hou2020} & ResNet50-FPN & 58.8 & \textbf{52.1} & \underline{63.7} & V100 & 99 \\
\midrule
\textbf{Ours} & RTFormer & \textbf{60.2} & \underline{51.5} & \textbf{66.5} & Titan RTX & \textbf{32} \\
\bottomrule
\end{tabular}
\end{center}
\vspace{-3mm}
\end{table*}
\begin{table}
\caption{Comparison of our method to state-of-the-art panoptic segmentation methods on the Mapillary Vistas validation set. The best results are shown in \textbf{boldface}, second best \underline{underlined}.}
\label{tab:mapillary}
\begin{center}
\begin{tabular}{lcccc}
\toprule
Method & Backbone & PQ $\uparrow$ & PQ$_{\text{th}}$ $\uparrow$ & PQ$_{\text{st}}$ $\uparrow$ \\
\midrule
Panoptic-DeepLab~\cite{Cheng2020} & ResNet50 & 33.3 & - & - \\
Seamless~\cite{Porzi2019} & ResNet50 & 36.2 & \underline{33.6} & 40.0 \\
Mask2Former~\cite{Cheng2021_1} & ResNet50 & 36.3 & - & - \\
Panoptic FCN~\cite{Li2021_2}  & ResNet50 & \underline{36.9} & 32.2 & 42.3 \\
EfficientPS~\cite{Mohan2021}  & - & \textbf{38.3} & \textbf{33.9} & \underline{44.2} \\
\midrule
\textbf{Ours} & RTFormer & 33.2 & 23.6 & \textbf{45.8} \\
\bottomrule
\end{tabular}
\end{center}
\vspace{-3mm}
\end{table}
We compare RT-K-Net with other real-time panoptic segmentation methods on the Cityscapes dataset in Table~\ref{tab:cityscapes}.
Compared to other methods, RT-K-Net sets new state-of-the-art performance on real-time panoptic segmentation with a PQ of $60.2~\%$ and an average runtime of $32~\si{ms}$.
\mbox{RT-K-Net} performs exceptionally well for stuff classes, where it outperforms all other methods by a large margin of $2.8~\%$ PQ$_{\text{st}}$.
We argue that this is mainly due to the strong baseline K-Net and the RTFormer backbone, which has a high representation ability for stuff classes.
For thing classes, RT-K-Net performs second best overall, being only $0.6~\%$ PQ$_{\text{th}}$ behind the method proposed by Hou~\etal~\cite{Hou2020}.
We find that RT-K-Net still struggles to discriminate large object instances due to crop-based training, which is a known problem in unified panoptic architectures~\cite{Geus2023}.
The runtime of all methods is evaluated on full resolution $1024 \times 2048$ pixel images.
For most methods, we use their reported runtime, based on a V100 GPU, which performs similarly or better than the Titan RTX used for RT-K-Net.
For MGNet~\cite{Schoen2021}, the paper reports runtime on an RTX 2080Ti GPU. 
Since the Titan RTX is much stronger, for a fair comparison, we use the official codebase of~\cite{Schoen2021} to compare their runtime to RT-K-Net.
Overall, RT-K-Net is the fastest method with an average runtime of $32~\si{ms}$.
The results show that RT-K-Net can push both accuracy and inference speed compared to other methods, making it directly applicable to real-time applications such as automated driving.
For Mapillary Vistas, we compare RT-K-Net to other state-of-the-art panoptic segmentation methods in Table~\ref{tab:mapillary}.
There are currently no real-time methods that provide results on Mapillary Vistas.
Hence, we compare RT-K-Net with state-of-the-art methods focusing on high accuracy rather than inference speed.
RT-K-Net shows promising results with a PQ of $33.2~\%$.
Again, RT-K-Net performs exceptionally well for stuff classes but struggles with thing classes, being $10.3~\%$ PQ$_{\text{th}}$ behind the method proposed by Mohan~\etal~\cite{Mohan2021}.
We argue that the higher discrepancy compared to Cityscapes lies in the higher amount of objects per image.
\mbox{Panoptic-DeepLab~\cite{Cheng2020}} is the only method that reports inference speed, taking on average $286~\si{ms}$ per image.
RT-K-Net matches the accuracy but is $75~\%$ faster with an average runtime of $69~\si{ms}$.
Qualitative results of RT-K-Net are shown in Fig.~\ref{fig:qualitative}.
The first two rows show results on the Cityscapes dataset, while the last two rows show results on the Mapillary Vistas dataset.
RT-K-Net performs well on Cityscapes and Mapillary Vistas, yielding accurate panoptic segmentation maps with crisp boundaries between predicted masks.
The last column shows error cases, where RT-K-Net either cannot find all object masks or struggles to predict masks of large objects like buses and trucks accurately.

\section{Conclusion}
In this work, we introduced RT-K-Net, the first unified architecture for real-time panoptic segmentation.
RT-K-Net is based on K-Net~\cite{Zhang2021} but addresses several shortcomings regarding real-time applications such as automated driving.
\mbox{RT-K-Net} introduces a novel mask normalization scheme, an optimized post-processing module, and an improved instance discrimination ability compared to K-Net.
Combined with a simplified mask and kernel initialization and low-latency backbone selection, RT-K-Net outperforms all previous real-time methods on the Cityscapes dataset.
In particular, our model reaches $60.2\%$~PQ with an average inference time of $32~\si{ms}$ for full resolution $1024 \times 2048$ pixel images on a single Titan RTX GPU.
We hope our work will inspire researchers further to investigate the real-time capability of panoptic segmentation methods.

\bibliographystyle{IEEEtran}
\bibliography{IEEEabrv,library}

\end{document}